%% file: full-paper-template.tex
\documentclass[pmlr]{jmlr}% new name PMLR (Proceedings of Machine Learning)

\RequirePackage{graphicx}
 \usepackage{booktabs}
\usepackage{longtable}% for long tables
\makeatletter
\def\set@curr@file#1{\def\@curr@file{#1}} %temp workaround for 2019 latex release
\makeatother
\usepackage[load-configurations=version-1]{siunitx} % newer version

\usepackage{graphicx}%
\usepackage{multirow}%
\usepackage{amsmath,amssymb,amsfonts}%
\usepackage{mathrsfs}%
\usepackage[title]{appendix}%
\usepackage{xcolor}%
\usepackage{textcomp}%
\usepackage{manyfoot}%
\usepackage{booktabs}%
\usepackage{algorithm}%
\usepackage{algorithmicx}%
\usepackage{algpseudocode}%
\usepackage{listings}%

\usepackage{adjustbox}
\usepackage{makecell}
\usepackage{subcaption}
\usepackage{xspace}
\usepackage{multicol}
\usepackage{threeparttable}
\usepackage{rotating}
\usepackage[defaultcolor=red]{changes}
\newcommand{\model}{REMed\xspace}
\usepackage{footmisc}

\theorembodyfont{\upshape}
\theoremheaderfont{\scshape}
\theorempostheader{:}
\theoremsep{\newline}

\jmlrvolume{252}
\jmlryear{2024}
\jmlrworkshop{Machine Learning for Healthcare}

\title[REMed]{\vspace{-1.5em}General-Purpose Retrieval-Enhanced Medical Prediction Model Using Near-Infinite History}

\author{\Name{Junu Kim}
       \Email{kjune0322@kaist.ac.kr}\\ 
       \addr Kim Jaechul Graduate School of AI\\
       KAIST\\
       Dajeon, Republic of Korea 
       \AND
       \Name{Chaeeun Shim}
       \Email{chaeeun@kaist.ac.kr}\\
       \addr Kim Jaechul Graduate School of AI\\
       KAIST\\
       Dajeon, Republic of Korea
       \AND
       \Name{Bosco Seong Kyu Yang}
       \Email{boscoskyang@outlook.com}\\
       \addr Department of Neurosurgery and Neurology\\
       Seoul National University Bundang Hospital\\
       Seongnam, Republic of Korea
       \AND
       \Name{Chami Im}
       \Email{chami0921@gmail.com}\\
       \addr Department of Surgery\\
       Seoul National University Bundang Hospital\\
       Seongnam, Republic of Korea
       \AND
       \Name{Sung Yoon Lim}
       \Email{nucleon727@gmail.com}\\
       \addr Department of Internal Medicine\\
       Seoul National University Bundang Hospital\\
       Seongnam, Republic of Korea
       \AND
       \Name{Han-Gil Jeong}\footnotemark[1]
       \Email{han.g.jeong@gmail.com}\\
       \addr Department of Neurosurgery and Neurology, Center for Artificial Intelligence in Healthcare\\
       Seoul National University Bundang Hospital\\
       Seongnam, Republic of Korea
       \AND
       \Name{Edward Choi}\footnotemark[1]
       \Email{edwardchoi@kaist.ac.kr}\\
       \addr Kim Jaechul Graduate School of AI\\
       KAIST\\
       Dajeon, Republic of Korea
       } 

\begin{document}

\maketitle

\let\oldthefootnote\thefootnote
\renewcommand{\thefootnote}{\fnsymbol{footnote}}
\renewcommand\thempfootnote{\fnsymbol{footnote}}
\footnotetext{\textsuperscript{*}Co-corresponding author}
\let\thefootnote\oldthefootnote

\vspace{-4em}
\begin{abstract}
Machine learning (ML) has recently shown promising results in medical predictions using electronic health records (EHRs).
However, since ML models typically have a limited capability in terms of input sizes, selecting specific medical events from EHRs for use as input is necessary.
This selection process, often relying on expert opinion, can cause bottlenecks in development.
We propose Retrieval-Enhanced Medical prediction model (\model) to address such challenges. 
\model can essentially evaluate unlimited medical events, select the relevant ones, and make predictions.
This allows for an unrestricted input size, eliminating the need for manual event selection.
We verified these properties through experiments involving 27 clinical prediction tasks across four independent cohorts, where \model outperformed the baselines.
Notably, we found that the preferences of \model align closely with those of medical experts. 
We expect our approach to significantly expedite the development of EHR prediction models by minimizing clinicians' need for manual involvement.
\end{abstract}

\section{Introduction}

A patient's medical records in a hospital are archived as a sequence of medical events (\textit{e.g.}, lab measurements, prescriptions, procedures) in electronic health records (EHRs). 
In recent years, machine learning (ML) has shown remarkable potential in predicting various medical outcomes (\textit{e.g.}, mortality, length of stay) using EHR data \citep{choi2016retain, rajkomar2018scalable, wang2020mimic}.
However, the sheer volume of events in EHRs presents a significant challenge for developing predictive models. 
For instance, a patient in an intensive care unit (ICU) typically generates thousands of events daily \citep{sanchez2018big}.
The computational requirements of ML models scale with the size of the input \citep{rumelhart1985learning, vaswani2017attention}, making it challenging to effectively harness all this information, even with efficient modern architectures specialized to handle long input \citep{choromanski2020rethinking, gu2021efficiently, ma2022mega}.

Accordingly, heuristic event selection is required to reduce the input size. 
This process typically relies on human decisions made by domain experts, such as experienced clinicians, which is costly and time-consuming.
This acts as a significant bottleneck in the model development process.
While some recent studies have explored methods to alleviate the need for event selection, none have addressed the issue of limited input size, a fundamental reason for such selection \citep{rajkomar2018scalable, deasy2020dynamic, steinberg2021language, nallabasannagari2020all, hur2023genhpf, hur2022unifying}.
As a result, none have completely eliminated the need for domain experts' involvement.
Therefore, our main objective is to develop a model capable of handling a near-infinite number of events, thereby eliminating this bottleneck.

Recent studies have explored methods to eliminate the need for feature selection \citep{rajkomar2018scalable, deasy2020dynamic, steinberg2021language, nallabasannagari2020all, hur2023genhpf, hur2022unifying}. 
Notably, empirical findings from some of these studies suggest that models incorporating more features often outperform those with selected features \citep{rajkomar2018scalable, nallabasannagari2020all, hur2022unifying, hur2023genhpf}.
In general, there are two dominant approaches to achieve this.
Formally, a medical event $e_i$ occurring at timestamp $t_i$ is typically composed of a medical code $c_i$ that provides high-level information (\textit{e.g.}, a medical code ``L123'' denotes ``Lab measure of white blood cells''), and accompanied details $d_i$ (\textit{e.g.}, ``Value=3.7'', ``Unit of Measurement=K/uL'', ``Flag=abnormal'').
The first way is mapping each $c_i$ and $d_i$ to a unique vocabulary \citep{rajkomar2018scalable, nallabasannagari2020all, deasy2020dynamic, steinberg2021language}.
However, given that a typical EHR contains tens of thousands of unique $c_i$ and $d_i$ \citep{johnson2016mimic, johnson2020mimic, pollard2018eicu}, this method often struggles to handle the rare ones. 
The second approach treats both the $c_i$ and $d_i$ as text, mapping them to a natural language space \citep{hur2022unifying, hur2023genhpf}.
This method ensures that $c_i$ and $d_i$ with similar meanings (e.g., the frequently occurred code ``Non-invasive blood pressure systolic'' and the less common ``Manual blood pressure systolic'') are represented similarly, often outperforming the first approach \citep{hur2022unifying, hur2023genhpf}.
Among them, GenHPF \citep{hur2023genhpf} achieved superior performance by utilizing all information of $d_i$.
However, none of these approaches have addressed the issue of limited input size, a fundamental reason for event selection.
As a result, they all rely on manual observation window selection to limit the number of input events to a computationally feasible scale.
This necessitates the involvement of domain experts, which becomes a significant bottleneck in the model development process.
Therefore, we aim to develop a model capable of handling a near-infinite number of events, thereby eliminating the need for feature and observation window selection.

\begin{figure}[t]%
\centering
\includegraphics[clip, width=1.0\textwidth]{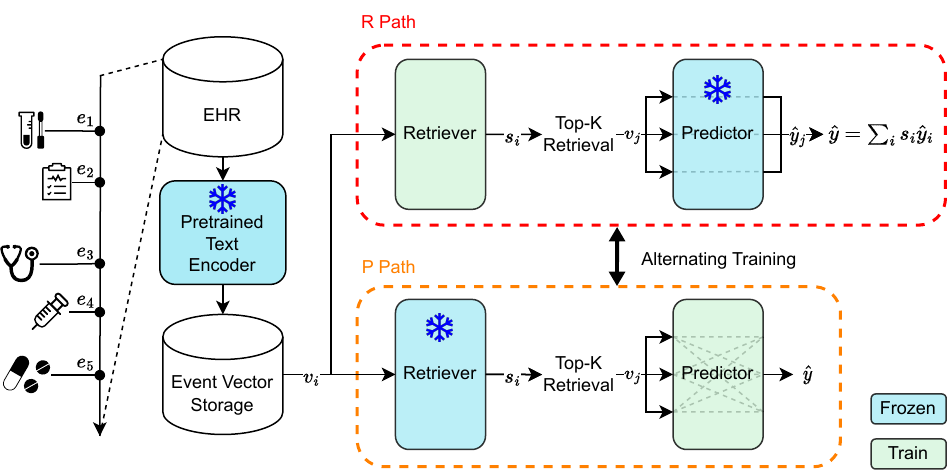}
\caption{Model Architecture: \model receives a series of event vectors as input, continuously identifies important events, retrieves them, and makes predictions. To ensure both Retriever and Predictor are trainable, our model alternates between two forward paths during the training stage. Note that the timestamps are omitted in this figure.}
\label{fig:model}
\end{figure}

We tackle this challenge by employing a Retrieval-Based Approach (RBA).
% with GenHPF \citep{hur2023genhpf}.
RBA, which has been widely explored in the natural language processing (NLP) question-answering (QA) domain, operates in two primary steps: 1) retrieving a collection of documents relevant to a specific question and 2) using these documents to make informed predictions \citep{karpukhin2020dense, lewis2020retrieval}. 
Inspired by RBA's capability to efficiently process millions of documents \citep{borgeaud2022improving}, we adopt its methodology for managing virtually infinite medical events. 
Our model, named \textit{Retrieval-Enhanced Medical prediction model} (\model), 1) retrieves events that are useful for predicting the target outcome, and 2) performs a prediction by leveraging the correlations among these selected events (Figure \ref{fig:model}). 
As a result, \model can process a near-infinite number of events\footnote{Further discussion about the near-infinite history is provided in \appendixref{complexity}.}, thereby eliminating the need for event selection, ultimately minimizing the domain expert involvement in the development process.

We trained \model on 27 clinical prediction tasks, including mortality, length of stay, creatinine, and platelets prediction, using four independent cohorts from publicly available EHR datasets: MIMIC-IV \citep{johnson2020mimic}, eICU \citep{pollard2018eicu}, UMCdb \citep{pollard2018eicu}, and HIRID \citep{hyland2020early}.
From this comprehensive evaluation, \model showcased its superior performance compared to various baselines. %, regardless of whether a short or long observation window was used.
Notably, \model's retrieval result is compatible with established medical knowledge.

In this study, we utilized open-source datasets only and made all our codes accessible to the public, guaranteeing transparency and reproducibility of our results.
In this study, we utilized open-source datasets only and made all our codes accessible to the public\footnote{\url{https://github.com/starmpcc/REMed}}, guaranteeing transparency and reproducibility of our results.
We believe that \model can accelerate the development of medical prediction models by minimizing the involvement of domain experts.

\subsection*{Generalizable Insights about Machine Learning in the Context of Healthcare}
 
A patient admitted to a hospital can generate millions of medical events in EHR, yet typical machine learning models possess limited capability in handling such extensive inputs. 
Therefore, selecting important medical events based on domain experts' knowledge was essential to develop a medical predictive model.
In this context, our contributions can be summarized as follows:

\begin{itemize}%[leftmargin=0.15in]
    \item We propose \model, the first attempt to introduce the Retrieval-Based Approach to the medical prediction task using structured EHR data.
    \model can handle virtually an unlimited number of events and demonstrates superior performance in handling a large number of events.
    \item \model eliminates the fundamental need for event selection due to its ability to manage unlimited events.
    We empirically demonstrated that abstaining from such selection does not compromise the prediction performance.
    \item \model can identify and retrieve clinically relevant events. 
    We verified that the retrieval results are compatible with established clinical knowledge.
\end{itemize}

\section{Backgrounds}

\subsection{Problem Definition}

Formally, a patient's medical history $H$ can be represented as:
\begin{equation}
    H = \{(e_1, t_1), (e_2, t_2), \ldots, (e_i, t_i), \ldots\},
\end{equation}
where $e_i$ is the $i^{th}$ medical event and $t_i$ is its corresponding timestamp.
Medical prediction aims to predict the specific outcomes of a patient (\textit{e.g.}, mortality) at a certain time-point using the patient's medical history, such that
\begin{equation}
    \hat{y} = f(\{(e_i, t_i) | t_i < T\}),
\end{equation}
where $f$ is a prediction model, and T denotes the moment the prediction is carried out (\textit{i.e.}, prediction time).

\subsection{Event Selection}
A medical event $e_i$ occurring at timestamp $t_i$ is typically composed of a medical code $c_i$ and accompanied details $d_i$.
There are two primary strategies in event selection:
1) Feature selection - This strategy focuses on selecting a particular set of $c_i$'s that are considered relevant to the prediction target;
2) Observation window selection - This strategy often selects recent events based on their $t_i$'s.
However, since both strategies rely on the heuristic decision of domain experts, it acts as a bottleneck in the model development process.

\subsection{Retrieval-Based Approach in Medical Domain}
A typical application of RBA in the medical domain is the retrieval of appropriate clinical articles \citep{simpson2014overview, conf/trec/RobertsSVH15, roberts2016overview, wei2018embedding}. 
Furthermore, some research has aimed to enhance medical prediction performance by using retrieved documents related to a given patient's diagnosis codes \citep{10.1145/3459637.3482273} or clinical notes \citep{naik-etal-2022-literature}. 
Additionally, another study has used RBA to retrieve clinical note snippets relevant to specific medical situations \citep{jiang2023conceptualizing}. 
However, all these works have focused on retrieving clinical text. 
In contrast, our work aims to retrieve medical events from structured EHRs that are related to a given medical prediction task.

% Cite 1: naik-etal-2022-literature: clinical note 주고 관련 paper retrieval -> mort pred
% Cite 2: 10.1145/3459637.3482273 EHR code 주고 paper retrieval -> health risk pred
% Cite 3: jiang2023conceptualizing 현재 의사에게 가장 필요한 note 가 뭔지 note를 retrieval
% Cite 4: wei2018embedding 노트 주고 웹텍스트 서치 -> 좀 애매함
% 일단 위에 세개 위주로 써 볼 것!

\section{Retrieval-Enhanced Medical Prediction Model}

This section explains our model architecture, as illustrated in Fig \ref{fig:model}. 
While conventional approaches necessitate feature or observation window selection to reduce the number of events, we aim to build a model without such conditions.
Consequently, as mentioned earlier, we start with GenHPF, which has demonstrated superior performance among the feature selection-free methods.
Following this, we first convert each event $e_i$ to its text representation $r_i$ by first converting the code $c_i$ to its description (e.g., ``L123'' $\rightarrow$ ``Lab measure for white blood cells'') and then concatenating with its accompanied details $d_i$ (e.g., ``Lab measure for white blood cells, Value 3.7, Unit of Measurement K/uL, Flag abnormal'').

Similar to the typical Retrieval-Based Approach (RBA) in question-answering (QA), each $r_i$ is initially encoded into a dense vector $v_i$ using a pre-trained text encoder $\textit{Enc}_\text{PT}$ \citep{alsentzer2019publicly, hur2023genhpf}.
\begin{equation}
    v_i = \textit{Enc}_\text{PT}(r_i)
\end{equation}

In a typical question-answering task, each document's importance is calculated by comparing each document to the given query, which has theoretically an infinite degree of freedom (\textit{i.e.}, a user can ask anything.). 
In contrast, for medical prediction tasks, typically a set of prediction targets (\textit{e.g.}, mortality, readmission) is fixed\footnote{One could try prompt-based medical prediction with a large language model (LLM), thus having unfixed prediction targets. 
% However, it is uncertain whether LLMs can perform medical predictions based on complex EHR. 
Further discussion is provided in \appendixref{apd:multitask}.}.
As a result, evaluating the events with respect to queries that vary for each prediction is not required.
Instead, we directly assess the scalar importance $s_i$ of each event vector $v_i$ while considering its timestamp $t_i$ using Retriever $R$, which is implemented with a multi-layer perceptron (MLP),
\begin{equation}
    s_i =  R(v_i, t_i).
\end{equation}
In this way, the information related to the prediction targets is embedded in the parameters of $R$.
Following this, the top-$k$ event vectors $v_j$, ranked by their scores $s_j$, are retrieved and fed into the Predictor $P$ along with their respective timestamps $t_j$, which is implemented with Transformer \citep{vaswani2017attention}.
$P$ interprets the meaning of events in relation to their surrounding events, making a prediction $\hat{y}$.
\begin{equation}\label{eq:pred}
    \hat{y} = P(\{v_j, t_j\})
\end{equation}
While processing all events simultaneously with a Transformer is impractical due to its high computational requirements, evaluating all events \textit{independently} with an MLP is feasible.
By limiting the input into the Transformer to only the most relevant events, we can harness the powerful predictive performance of the Transformer while also ensuring computational efficiency.

Our training objectives are twofold: To train $R$ to understand the significance of each event, and to train $P$ to exploit the correlations among events.
It is, however, not straightforward to train $R$ and $P$ in an end-to-end manner, which requires that $s_j$ directly affect $\hat{y}$ while acting as an event importance indicator.
Feeding $s_j$ into equation \ref{eq:pred} will only partially satisfy this requirement (see \appendixref{apd:path} for further discussion on this point), and therefore we devise a new training strategy that involves two paths, namely the \textit{$R$ Path}, and the \textit{$P$ Path}.
Each path exclusively trains one component, $R$ or $P$, while keeping the other component frozen. 
Throughout the training process, we alternate between these two paths at each step. 
In the $R$ Path (Figure \ref{fig:model}, red box), we feed each event to the frozen $P$ independently, making the same number of predictions as the number of retrieved events. 
These predictions are then combined based on their importance scores to make a final prediction.
\begin{equation}\label{eq:ret}
    \hat{y} = \sum _j s_j \hat{y}_j, \enskip \mbox{where} \enskip \hat{y}_j = P(v_j, t_j)
\end{equation}
Following this, $s_j$ is directly affecting $\hat{y}$ while acting as an event importance indicator, since $R$ would be trained to increase $s_j$ when $\hat{y}_j$ is consistent with the label $y$.
Therefore, $R$ can learn to calculate the importance of each event.
It might seem possible to train $P$ using the $R$ Path; however, this cannot train $P$ effectively.
Since each event is fed into $P$ independently, $P$ cannot learn the correlation between events using the $R$ Path alone. 
Therefore, in the $P$ Path (Figure \ref{fig:model}, orange box), we feed all retrieved events from the frozen $R$ into $P$ simultaneously, thereby training $P$ to make predictions considering the correlation between events (\eqref{eq:ret}).
During the evaluation, REMed relies solely on the $P$ Path, drawing from the combined strengths of both $R$ and $P$.

% In the $R$ Path, we make predictions using each event independently, then combine them based on their importance score to make a final prediction.
% \begin{equation}\label{eq:ret}
%     \hat{y} = \sum _j s_j \hat{y}_j, \enskip \mbox{where} \enskip \hat{y}_j = P(v_j, t_j)
% \end{equation}
% Following this, $s_j$ is directly affecting $\hat{y}$ while acting as an event importance indicator, since $R$ would be trained to increase $s_j$ when $\hat{y}_j$ is consistent with the label $y$.
% Therefore, $R$ can learn to calculate the importance of each event.
% However, the $R$ Path alone can cause $P$ to be biased towards making predictions based on individual events, which is far from our intention.
% Therefore, in the $P$ Path, we train $P$ based on all retrieved events (\textit{i.e.}, equation \ref{eq:pred}), so that it can consider correlations among multiple events.
% Throughout the training process, we alternate between these two paths.
% Note that in the $R$ Path, we only update the parameters of $R$ while keeping $P$ frozen. 
% In the $P$ Path, $R$ is naturally frozen as $s_j$ and does not take part in equation \ref{eq:pred}.
% During the evaluation, \model relies solely on the $P$ Path, drawing from the combined strengths of both $R$ and $P$.

In this manner, we successfully built a powerful prediction model that can process a near-infinite unlimited number of events. 
By resolving the fundamental cause of the event selection, \model significantly reduces the need for a domain expert's involvement, ultimately resolving a significant bottleneck in the model-building process.
For additional details, please refer to \appendixref{apd:model}.

\section{Cohorts and Experimental Settings}

Even if \model can bypass event selection, its practicality will be limited if this bypass decreases its prediction performance. 
While some research suggests that abstaining from feature selection does not compromise performance \citep{rajkomar2018scalable, nallabasannagari2020all, hur2022unifying, hur2023genhpf}, there are no such results for observation window selection. 
Accordingly, we aim to demonstrate the following two key properties of \model:
1) \model can effectively handle long inputs compared to multiple baselines, and 
2) the performance of \model is not compromised when the observation window selection is bypassed.
We validated these properties through extensive experiments using four publicly available EHR datasets: MIMIC-IV \citep{johnson2020mimic}, eICU \citep{pollard2018eicu}, UMCdb \citep{thoral2021sharing}, and HIRID \citep{hyland2020early}.
These datasets are commonly employed in medical prediction research \citep{mcdermott2021comprehensive, hur2022unifying, hur2023genhpf}, and their wide accessibility guarantees the reproducibility of our experiments by the research community. 
Furthermore, these datasets consist of EHRs from ICU-admitted patients, meaning the events are densely recorded (\textit{i.e.}, a large number of medical events).
This characteristic is advantageous for showcasing \model's strength in processing long inputs.
We minimally filtered these datasets based on two criteria: patients who were adults (age $>18$) and those with an ICU stay exceeding 48 hours.
Detailed experimental setups can be found in \appendixref{apd:setting}.

\input{tables/task}

We demonstrated \model's robust capabilities under various conditions.
First, we examined \model and compared it with baselines using four datasets.
Second, we tested our model at two prediction times: 24 hours and 48 hours after ICU admission. 
Third, we trained and evaluated our model on ten categories and 27 tailored medical prediction tasks (Table \ref{tab:task}), in a multi-task manner\footnote{While examining \model for each task can also showcase its robustness, building multiple models corresponding to each task comes with severe overhead in practical scenarios. 
Further discussion is provided in \appendixref{apd:multitask}.
}.
In addition to the administrative prediction tasks commonly used in prior research \citep{rajkomar2018scalable, wang2020mimic, mcdermott2021comprehensive, steinberg2021language, hur2022unifying, hur2023genhpf}, we further added frequent lab measurement prediction tasks that are closely related to a patient's overall status.

\begin{figure}[!ht]
    \centering
    \includegraphics[clip, width=1.0\textwidth]{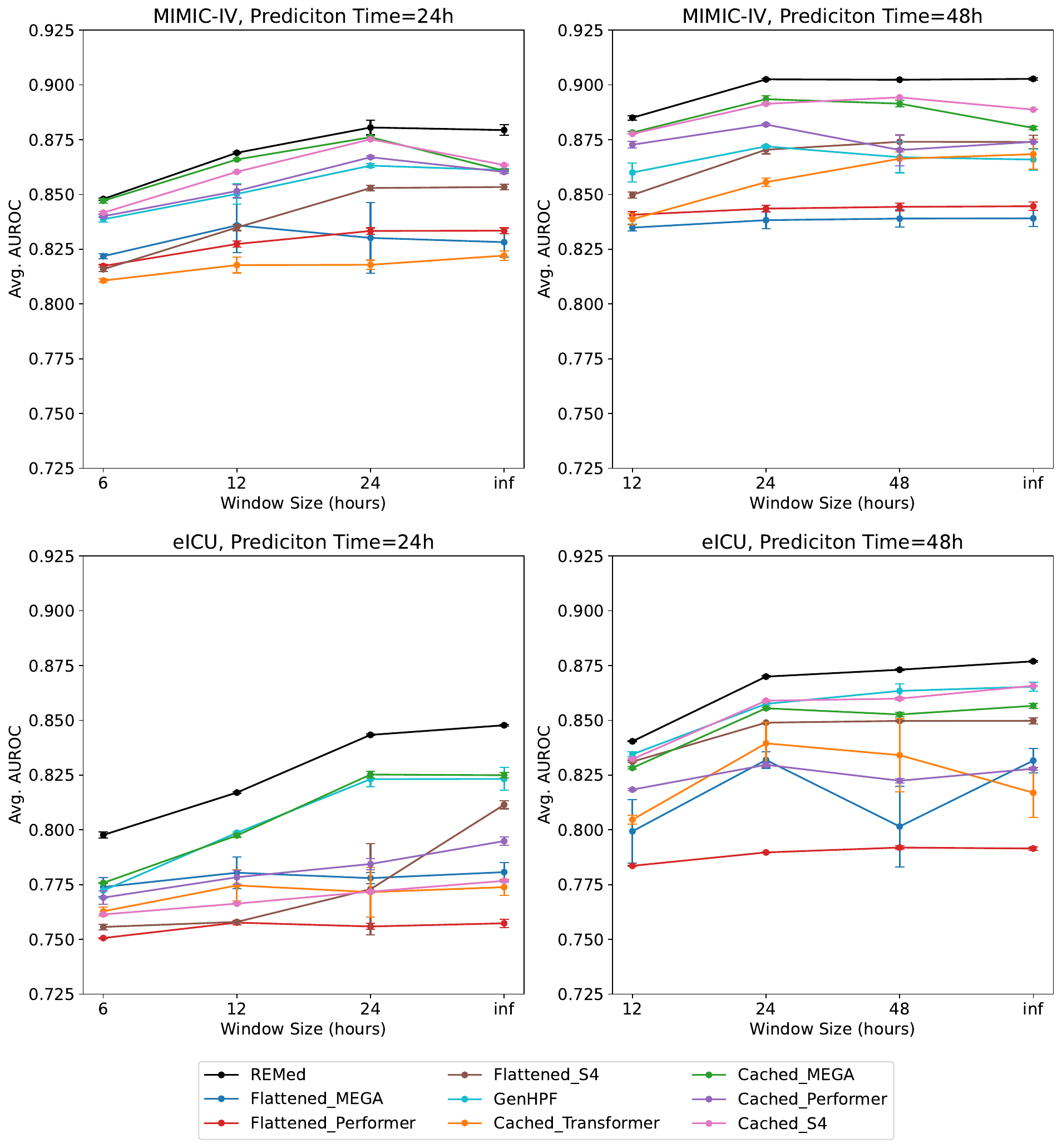}
    \caption{Performance Analysis Result on MIMIC-IV and eICU. We evaluated \model and the baselines on two datasets, two prediction times, and multiple observation window sizes. The y-axis corresponds to the micro-average AUROC over tasks. The error bars represent the standard error of the mean for three runs with different random seeds.}
    \label{fig:main_plot}
\end{figure}

We compared \model with various baselines, including GenHPF \citep{hur2023genhpf} and its variants.
\textit{GenHPF}, the basis of our model, uses two Transformer \citep{vaswani2017attention} models in an end-to-end manner, one for encoding each medical event into a vector representation and another for making predictions.
\textit{Flattened} model, a variant proposed in the same paper \citep{hur2023genhpf}, concatenates all $r_i$ in chronological order, and passes them to a single Transformer model.
Additionally, we introduce \textit{Cached} model, which uses the event vectors $v_i$ as input to a single Transformer model, similar to \model.
Since all these baselines can only process a limited number of events, we prioritized the most recent events as input when the input size reaches the computational limit.
Additionally, to partially alleviate this input size restriction, we replaced their backbone Transformer with modern, efficient architectures that are specialized for processing long inputs.
Specifically, we selected Performer \citep{choromanski2020rethinking}, S4 \citep{gu2021efficiently}, and MEGA \citep{ma2022mega}, which have demonstrated state-of-the-art performance in the benchmark for long input \citep{tay2020long}.
This modification enables the baselines to handle a larger number of events.
We also considered RMT \citep{bulatov2022recurrent} as a backbone; however, it failed to converge without the specific curriculum learning they proposed (\appendixref{apd:rmt}). 
Details about these baselines can be found in \appendixref{apd:baseline}.

\section{Results}\label{results} 

\subsection{Performance Analysis}

We primarily focus our experiments on MIMIC-IV and eICU, and further validate our findings using UMCdb and HIRID.
The results are displayed in Figure \ref{fig:main_plot}.
First, \model outperformed all baseline models in most settings, regardless of whether the long or short observation window was used.
To statistically affirm the superior prediction performance of \model, we employed a one-sided Mann-Whitney U test \citep{mann1947test} on each dataset, prediction time, and observation window size, comparing it against the best baseline performance at each setting.
The results substantiated \model's superiority in all settings ($p<0.05$), barring two cases (MIMIC-IV, prediction time 24h, observation windows 6h and 24h).
Even for those two cases, our model's performance was still on par with the best baselines.
Therefore, we can conclude that \model processes long input more effectively than the baselines.

\begin{figure*}[!t]
    \centering
    \includegraphics[clip, width=1.0\textwidth]{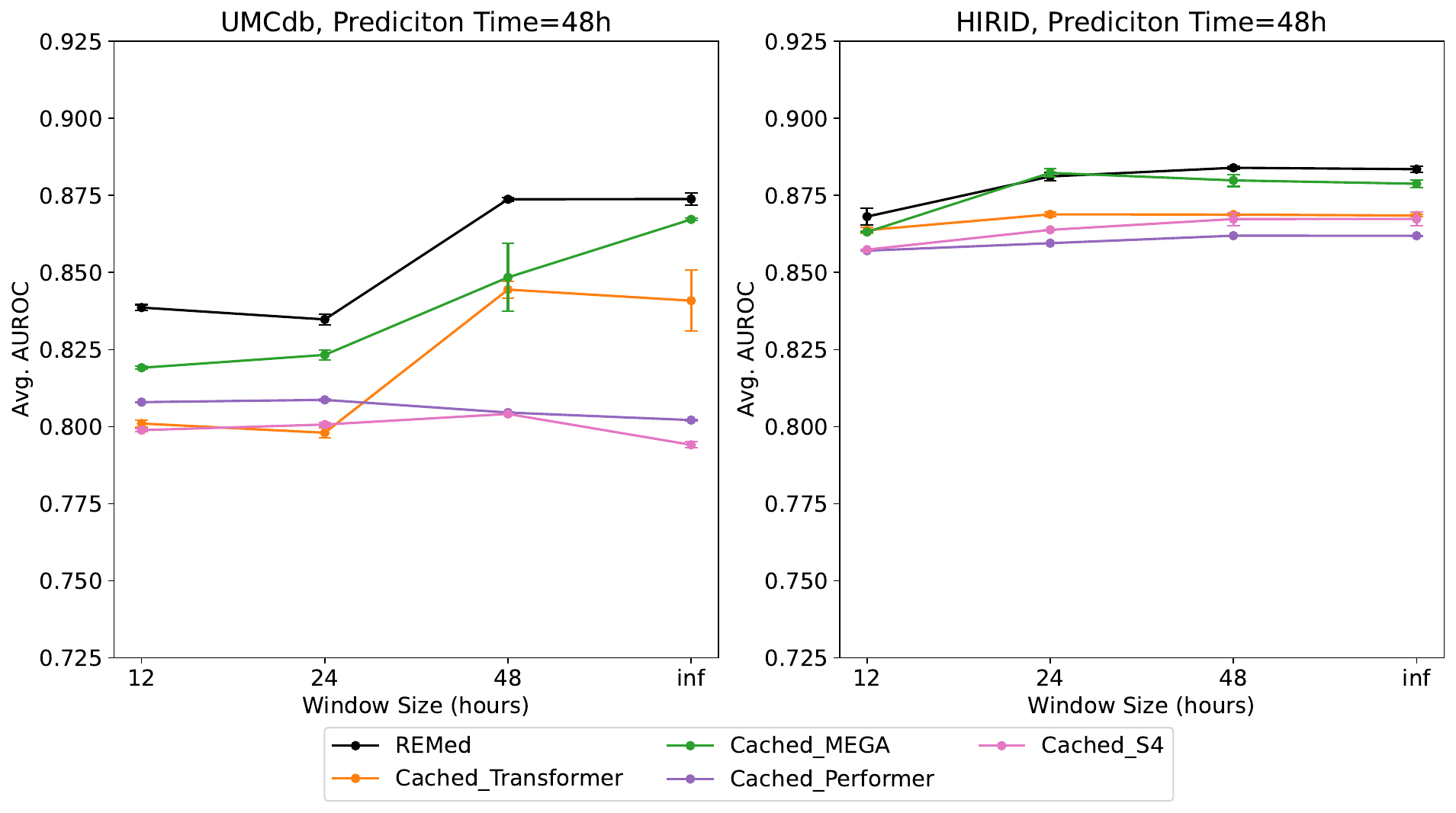}
    \caption{Performance Analysis Result on UMCdb and HIRID. We evaluated \model and the baselines on  multiple observation window sizes. The y-axis corresponds to the micro-average AUROC over tasks. The error bars represent the standard error of the mean for three runs with different random seeds.}
    \label{fig:additional}
\end{figure*}

We noticed a monotonic increase in our model's prediction performance as the observation window length was extended. 
This is supported by the Kendall-Tau test \citep{kendall1938new}, with $p$-values ranging from 0.001 to 0.01 across all four graphs. 
Even though the performance plateaued in MIMIC-IV, no decrease was observed with longer windows.
In contrast, the performance of most baseline models either plateaued or declined.
Even for those that occasionally demonstrated monotonically increasing performance, their results were inconsistent across the four settings.
We conclude that bypassing the observation window selection with \model does not compromise the performance; the unlimited observation window consistently yields the best performance.
We believe it is all the more valuable that \model outperformed the baselines in the unlimited observation window scenarios.

To more rigorously demonstrate the two aforementioned properties, we expanded our experiments to include UMCdb and HIRID.
To simplify the experiments, we focused on the 48-hour prediction window, which allows for longer input lengths.
Furthermore, we excluded the flattened baselines, which consistently showed inferior performance in earlier experiments.
The results of these experiments are illustrated in Figure \ref{fig:additional}.
Following the same analytical approach as above, the Mann-Whitney U test validated the superiority of our model over the baselines with a significance level of p$<$0.05 in most cases. 
There were two exceptions for shorter window sizes in HIRID (12h and 24h); however, even in such cases, our model exhibited comparable performance to the best baseline.
Similarly, the Kendall-Tau test verified that the prediction performance of our model monotonically increased as the observation window length increased for each dataset, with significance levels of p$<$0.05 and p$<$0.01.

We conducted additional experiments to validate REMed's robustness and strong performance.
First, we analyzed the per-task performance from the previous experiments and confirmed that REMed consistently outperformed the baselines in most scenarios, demonstrating its task-wise generalizability (see \appendixref{apd:per_task}).
Second, we repeated the previous experiment with a different model configuration and verified that the two aforementioned properties were maintained (see \appendixref{apd:small}). 
Lastly, we established that REMed provides a significant performance advantage over traditional regression models, despite its complexity (see \appendixref{apd:regression}).
% REMed demonstrates strong performance across different model sizes (\appendixref{apd:small}), various tasks (\appendixref{apd:per_task}), and in comparison to regression models (\appendixref{apd:regression}).
% Additionally, we conducted these analyses under a different model configuration and found that these two properties still held (\appendixref{apd:small}).
% For a per-task performance analysis, REMed outperformed the baselines in most of tasks, validated its generalizability over tasks (\appendixref{apd:per_task}).
Owing to this robust and powerful capability to process near-infinite events, \model can minimize the need for manual involvement of domain experts, a common bottleneck in developing medical prediction models.

\subsection{Retrieval Result Analysis}

\input{tables/topk}

While retrieval-based models in the general domain measure their retrieval performance using labeled data, \citep{karpukhin2020dense, lewis2020retrieval}, there is no labeled data in medical prediction tasks (\textit{i.e.}, ground-truth label indicating which event(s) must be retrieved).
Therefore, we indirectly measured the retrieval performance of the Retriever $R$ by analyzing whether its behavior is compatible with established clinical knowledge.
In this section, we exclusively focused on MIMIC-IV and eICU.

We first checked which medical codes $c_i$ were frequently retrieved.
We calculated the average number of events corresponding to a specific code retrieved every time the model makes a prediction for all test set samples.
We narrowed our analysis to the 250 medical codes that occurred most frequently in the test set and used our best-performing models for each cohort.
These best models were trained on a 48-hour prediction time and an unlimited observation window.
Table \ref{tab:topk} displays the top-30 most frequently retrieved codes from each cohort.
Our model frequently retrieved codes related to core lab measurements, vital signs, neurologic status, analgosedative drugs, ventilation data, and input/output records.

To check whether these codes were truly useful for predicting the target tasks, we conducted an expert test involving two professors and a clinical fellow, all with expertise in the ICU.
For each dataset, we showed them the same 250 codes and asked them to identify the 30 most significant ones.
The average overlap of the top-30 codes between any two clinicians was 12.8 out of 30.
While this overlap value might seem low, they do not necessarily indicate a lack of agreement.
In fact, this reflects the complex nature of clinical decision-making, where multiple valid perspectives can exist.
Clinicians may prioritize different codes based on their unique experiences and specialties, leading to a degree of variability in the selection.
The degree of overlap between the top 30 codes of our model and each clinician's selection averaged 10.9 out of 30. 
Although the agreement between the model and clinicians was slightly lower than that between two clinicians, this discrepancy may be due to inherent differences between models and human judgment. 
For instance, when both a high-level and low-level code (\textit{e.g.}, a chart event code \textit{heart rate alert} and a vital sign code \textit{heart rate}) are available, clinicians tend to prefer the former, while \model the latter.
Given this, the alignment of our model's choices with those of the clinicians was roughly equivalent to the alignment observed between different clinicians.
This suggests that $R$ can identify codes useful for the target task.

\begin{figure*}[!t]

\begin{minipage}[c]{.5\linewidth}
    MIMIC-IV
    \centering
\end{minipage}%
\begin{minipage}[c]{.5\linewidth}
    eICU
    \centering
\end{minipage}

\vspace{0.5em}

\raisebox{-0.5\height}{(a)}
\begin{minipage}[c]{.5\linewidth}
\centering
\includegraphics[width=1.0\textwidth]{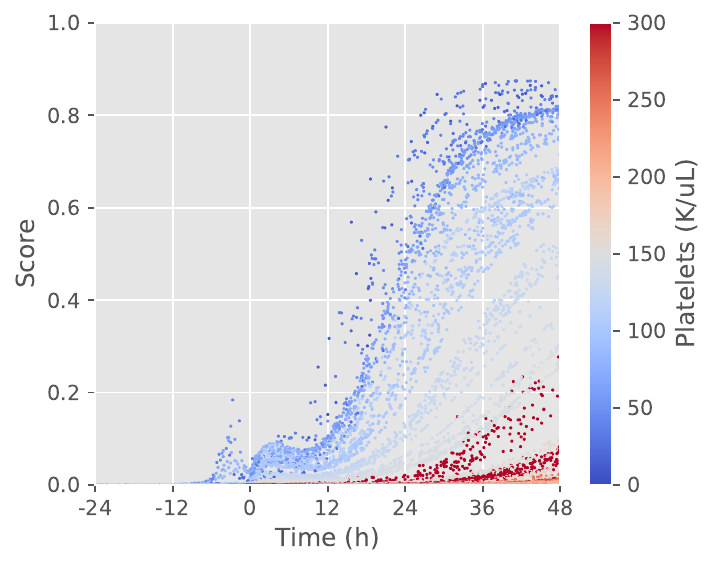}
\end{minipage}
\begin{minipage}[c]{.5\linewidth}
\centering
\includegraphics[width=1.0\textwidth]{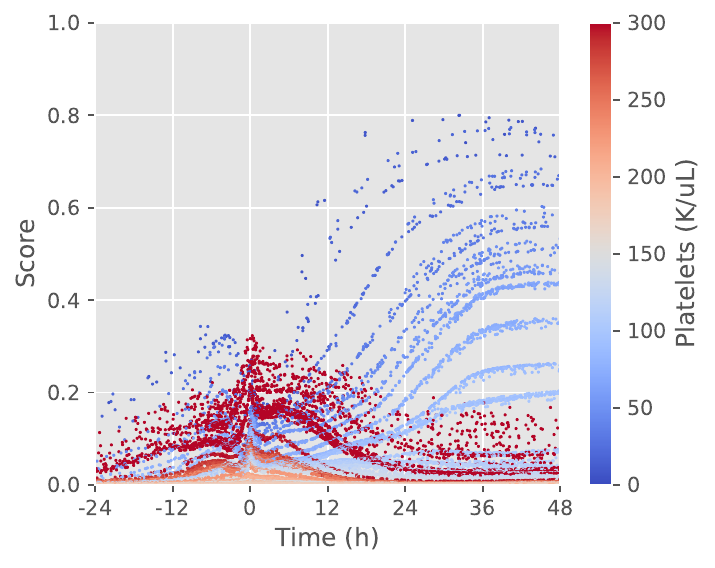}
\end{minipage}

\raisebox{-0.5\height}{(b)}
\begin{minipage}[c]{.5\linewidth}
\centering
\setlength{\tabcolsep}{2pt}
\footnotesize{
\begin{tabular}{lrrr}\toprule
Timestamp &Itemid &Value \\\cmidrule{1-3}
123 &Heart Rate &103 \\\cmidrule{1-3}
123 &Respiratory Rate &22 \\\cmidrule{1-3}
\vdots &\vdots &\vdots \\\cmidrule{1-3}
123 &Blood Pressure Diastolic &102 \\\midrule
\bottomrule
\end{tabular}
}
\end{minipage}%
\begin{minipage}[c]{.5\linewidth}
\centering
\setlength{\tabcolsep}{2pt}
\footnotesize{
\centering
\begin{tabular}{lrrrrr}\toprule
Timestamp &heartRate &respiration &$\cdots$ &diastolic \\\cmidrule{1-5}
123 &103 &22 &$\cdots$ &102 \\\midrule
\bottomrule
\end{tabular}
}
\end{minipage}

\caption{Retriever Analysis Result. The left column is for MIMIC-IV, and the right column is for eICU. (a) Allocated importance scores of platelets events against their timestamps and ``value'' fields. (b) Example of vital events from both datasets.
}
\label{fig:analysis}
\end{figure*}

In addition to analyzing the medical code $c_i$, we explored the effects of the accompanied details $d_i$ and the timestamp $t_i$ on the behavior of the Retriever $R$.
$d_i$ is composed of various fields (\textit{e.g.}, value, unit of measurement, flag, comment), and the composition varies based on the category of events, such as lab measurements or prescriptions, complicating the analysis.
We specifically focused on lab measurement events associated with a platelets code, allowing us to clarify the fields of $d_i$. 
Typically, $d_i$ for a lab measurement event includes ``value'', ``unit of measurement'', and ``flag'' fields. 
The ``unit of measurement'' remains consistent for the same code, and the ``flag'' is often derived from the ``value''.
Hence, we analyzed the scalar importance $s_i$ in relation to $t_i$ and the ``value'' for events associated with the lab codes used as prediction targets.
Figure \ref{fig:analysis} (a) presents the results for the platelet measurement code, while the results for other codes are displayed in \appendixref{apd:importance}.
$R$ assigned high scores to recent events or those with values in the abnormal range.
These events are also regarded as important based on clinical knowledge.
While peaks are observed around the ICU admission time, especially in the case of eICU, the lab results at this moment are typically regarded as pivotal in predicting future outcomes \citep{ferreira2001serial}.

From the analysis presented above, the overall trends in both datasets are similar, but there are a few notable differences.
These variations can be attributed to the unique characteristics inherent in each dataset.
For example, the ``vitalPeriodic'' code is frequently retrieved from eICU as seen in Table \ref{tab:topk}, whereas no single dominant code exists in MIMIC-IV.
In the eICU EHR system, 16 types of vital signs—such as respiratory rate, heart rate, and blood pressure—are consolidated under a single event with the code ``VitalPeriodic'' (Figure \ref{fig:analysis} (b, right)). 
On the other hand, in MIMIC-IV, each vital sign is recorded as a separate event with its unique code (Figure \ref{fig:analysis} (b, left)). 
This leads to the frequent retrieval of events with the ``vitalPeriodic'' code in eICU, while in MIMIC-IV, events associated with various vital sign codes are retrieved more evenly.
This behavior not only aligns with established clinical knowledge but also suggests \model's potential adaptability across different datasets.

In conclusion, the Retriever $R$ can correctly identify useful events for predicting the target tasks, based on $c_i$, $d_i$, and $t_i$, and its behavior was compatible with established clinical knowledge.
Additionally, \model showcased its adaptability to various characteristics of datasets.

\section{Discussion}\label{Discussion}

Utilizing structured EHR for medical prediction has been limited by the sheer volume of events, making the modeling process heavily reliant on expert opinions.
By introducing a Retrieval-Based Approach (RBA) to medical prediction, we demonstrated that \model can process a near-infinite number of events, thereby reducing the need for domain expert involvement.

Besides, REMed also outperformed the baseline models in most settings.
This superior performance can be analyzed by comparing it with the Cached Transformer since REMed's Predictor uses a Transformer architecture.
When input sizes exceed the Cached Transformer’s capacity, any performance gains can be attributed to the Retriever, which selects important past events that the Cached Transformer cannot observe. 
Conversely, when input sizes are shorter than the Cached Transformer's capacity, the Cached Transformer can observe all events and is expected to perform equally well or better than REMed. 
However, REMed still outperforms it, indicating that the Retriever might help by filtering out noisy events and passing only the important ones to the Predictor, possibly acting as an additional regularizer.
Further research is needed to clarify these observations.

We emphasize that our research solely relies on publicly available datasets and make our code publicly accessible for transparency and reproducibility.
We believe that \model can expedite the development of medical prediction models by reducing the dependency on domain experts.

\subsection*{Limitations and Future Works}
One of the major limitations of \model is its inability to account for the correlation between events when evaluating the importance score $s_i$. 
Although the Predictor $P$ can partially mitigate this limitation by considering the correlations in its predictions, this may not be sufficient for complex tasks where understanding the relationship between events is vital.

Another limitation of \model is its need for retraining to adapt to new tasks, but there is room for improvement.
Recently, the zero-shot \citep{wei2021finetuned} and few-shot \citep{brown2020language} capabilities of large language models (LLMs) have been demonstrated for general domain tasks.
This suggests that integrating LLMs with \model might be able to lessen the burden of additional training for new tasks.
However, smaller, supervised models often outperform LLMs on specific tasks since LLMs are primarily designed to predict the next natural language tokens autoregressively \citep{brown2020language, wei2021finetuned}. 
Therefore, integrating medical prediction with LLMs remains a challenging task. 
Nevertheless, considering the rapid development of LLMs, it is worth exploring their potential for such applications.

While our framework primarily focuses on structured EHR data, it can be extendable to other modalities. 
For instance, since we use a pretrained text encoder to encode events, adapting our framework to handle unstructured EHRs (e.g., clinical notes) would be straightforward. 
Incorporating other modalities, such as chest X-ray images, would be more challenging as it would require additional modality-specific encoders, but it is feasible.
Further work will be necessary to explore these potential extensions.

% ACKNOWLEDGEMENTS ONLY GO IN THE CAMERA-READY, NOT THE SUBMISSION
\acks{
This work was supported by the SNUBH-KAIST Joint Graduate Research Project on AI, the Institute of Information \& Communications Technology Planning \& Evaluation (IITP) grant (No.RS-2019-II190075), National Research Foundation of Korea (NRF) grant (NRF-2020H1D3A2A03100945), and Korea Medical Device Development Fund grant (Project Number: 1711138160, KMDF\_PR\_20200901\_0097), funded by the Korea government (MSIT, MOTIE, MOHW, MFDS), Cloud TPUs from Google's TPU Research Cloud (TRC).
}

%Do NOT change font size of references or modify the bibliography style
\bibliography{sample}

\newpage
\appendix

\section{Model Design}\label{apd:model}
In this section, we describe the design choices of \model. 
Unless specifically mentioned, all experiments described in this section are performed using the MIMIC-IV dataset, a 48-hour prediction time, and an unlimited observation window setting.
All of the experiments were performed with a single A6000 48G GPU.

\subsection{Event Encoding}
We investigated both unsupervised and supervised models for the event encoder $Enc_{\text{PT}}$, which encodes each event $e_i$ into a vector $v_i$.
First, we employed Bio-ClinicalBERT \citep{alsentzer2019publicly}, a derivative of BERT, further unsupervised pre-trained on biomedical and clinical domain literature.
Despite its widespread use for clinical text encoding, it has been trained with MIMIC-III clinical notes, presenting two issues. 
1) Since MIMIC-III and MIMIC-IV have overlapping patients, some of the samples in our test set might have been exposed during the training of the model.
2) There is a potential discrepancy in the distribution between note data and event data.
To evaluate an unsupervised event encoder without these issues, we additionally trained a Transformer \citep{vaswani2017attention} from scratch with a masked language modeling objective.
This model is trained using the text representations $r_i$'s as input, which originated from the training set of our MIMIC-IV cohort.
Lastly, we used GenHPF \citep{hur2023genhpf}, a supervised medical prediction model that employs two Transformers for event encoding and prediction.
For our purposes, we trained GenHPF and employed the first Transformer as the event encoder.

We trained \model using the $v_i$'s encoded by these models, respectively. 
From the preliminary evaluation, the version of \model using $v_i$'s encoded with GenHPF achieved the best AUROC of 0.8747, compared to Bio-ClinicalBERT (0.7901) and MLM-based approach (0.8456).

However, GenHPF is primarily designed to predict based on a limited number of recent events. 
As a result, it struggles to encode events that occurred far back in a patient's history, such as emergency department events.
To mitigate this, we randomly sampled events from the patient's entire history and fed them as input, thereby achieving an AUROC 0.9027 (top-right of Figure \ref{fig:main_plot}).
We adopted this modified version of the GenHPF event encoder in all other experiments presented in this paper.

\subsection{Importance Scoring}
In the Retrieval-Based Approach, both the question given by the user and documents are encoded as vectors. The cosine similarity between these vectors is then computed to determine the relevance of the documents to the question.
To adapt this methodology for medical prediction, one might consider using a trainable vector that represents the predefined task (\textit{e.g.}, mortality prediction) and then measuring its cosine similarity with the event vectors.
We preliminarily compared the cosine similarity method with the Multi-Layer Perceptron (MLP) method.
For simplicity, we ignored $t_i$ and used $v_i$ exclusively as input in this comparison.
The results indicated that the MLP method outperformed the cosine similarity method, with scores of 0.8898 versus 0.8849. 
%However, in our experiments, employing a Multi-layer Perceptron (MLP) yielded better performance compared to the straightforward cosine similarity method (0.8898 vs. 0.8849).
Moreover, we encountered challenges when trying to incorporate temporal information, which can significantly affect the performance, into the cosine similarity method.
In contrast, when we concatenated $v_i$ and $t_i$ and input them into the MLP, there was a noticeable performance improvement, reaching an average AUROC of 0.9027.
Using this scalar importance $s_i$, \model retrieves the top-$k$ event vectors $v_i$.
Empirical testing on the validation set, with $k$ values ranging from 64 to 1024, revealed that setting $k$ to 128 consistently delivered the best performance.

\subsection{Training Path}\label{apd:path}

Using $s_i$ to retrieve the top-$k$ documents cannot make the gradients reach the Retriever $R$.
Hence, $s_i$ must be directly involved in the final prediction to render $R$ trainable. 
Furthermore, $s_i$ must indicate the event's importance to be used for top-$k$ retrieval.
Incorporating $s_i$ naively into equation \ref{eq:pred} can satisfy the first requirement. 
This integration enables backpropagation from the prediction loss to $R$, allowing both $R$ and $P$ to be trainable end-to-end.
However, because the top-$k$ retrieval operation does not propagate the gradient, $P$ cannot recognize that $s_j$ should reflect the importance of events, thus failing to meet the second requirement.

In contrast, our proposed method, \textit{R} Path, effectively addresses both of these challenges.
While $s_i$ is directly involved in the final prediction, $R$ is trained to increase $s_i$ when $\hat{y}_i$ is consistent with $y$.
The $s_i$ trained in this manner signifies the event's importance, and can therefore be used for top-$k$ retrieval.

\subsection{Multi-Task Prediction}\label{apd:multitask}
As previously mentioned, though training and evaluating medical prediction models for each prediction task is possible, this approach is impractical in real-world scenarios. 
The overhead involved in developing and operating numerous models makes using a single, multi-task model a more pragmatic choice.
Thus, we evaluated our model in a multi-task setting to validate its robustness across various tasks and its practicality in real-world scenarios. 
For comparison, we also provide the model's performance in a single-task setting. 
When we trained \model for each task and averaged the AUROC, it yielded 0.8978.
In contrast, the multi-task version of the model achieved 0.9027.
% This improvement can be attributed to the similarities among our target tasks, which resulted in a synergistic effect.

\subsection{Model Complexity and Near-Infinite History}\label{complexity}
\model consists of an MLP Retriever $R$ and a Transformer Predictor $P$.
$R$ evaluates each event vector independently, and each evaluation demands a constant amount of computation and memory. 
This means processing a patient's history with $R$ is linear in computational requirements relative to the number of events. 
On the other hand, although Transformer demands quadratic computational resources based on the input size \citep{vaswani2017attention}, $P$ always receives a fixed number of event vectors $v_i$, ensuring constant computational needs.
Hence, \model achieves linear complexity in computation and memory with the number of events, making it even more efficient than the contemporary architectures \citep{choromanski2020rethinking, gu2021efficiently, ma2022mega}.

Under the finite observation windows, \model's memory consumption remained below 2GB.
During the training of \model with the longest hospital stay record, which consisted of 267k ($\sim2^{18}$) medical events corresponding to about 400 hospital days and 85 ICU days, the peak memory usage was roughly 37GB.
In evaluation mode, \model processed up to $2^{20}$ dummy events within the memory constraints of an A6000 48G GPU.
Since this is four times longer than the extremest case among the four datasets, we claim that our model can process a near-infinite record of a single hospital admission.

\section{Experimental Detail}\label{apd:setting}

To maximize data utilization, we applied minimal filtering to our datasets: patients had to be over 18 years of age, and their ICU stays needed to exceed 48 hours.
Additionally, we treated each ICU admission within a single hospital stay as a separate model input.
For instance, under the 48-hour prediction time setting, if a patient was admitted to the ICU twice during a single hospital stay and each ICU stay exceeded 48 hours, we generated two separate model inputs.
The first input spanned from the time of hospital admission to 48 hours after the first ICU admission.
The second input spanned from the time of hospital admission to 48 hours after the second ICU admission, including the duration of the first ICU stay. 
We divided the cohorts into an 8:1:1 ratio for training, validation, and test sets. 
We also ensured that all ICU stays from a single patient were grouped into the same partition to prevent potential test set leakage.
The statistics and label distribution for the datasets are provided in Tables 3-8.

\input{tables/length}
\input{tables/stat}
\input{tables/label}

For MIMIC-IV, we used the following tables: hosp/labevents, hosp/prescriptions, hosp/microbiologyevents, icu/inputevents, icu/chartevents, icu/outputevents, icu/procedureevents, ed/medrecon, ed/pyxis, ed/vitalsign, ed/diagnosis, and ed/triage. 
For eICU, we used the following tables: lab, medication, microLab, infusionDrug, intakeOutput, nurseCharting, nurseCare, nurseAssessment, treatment, vitalAperiodic, and vitalPeriodic. 
For UMCdb, we used the following tables: drugitems, freetextitems, listitems, numericitems, procedureorderitems, processitems.
For HIRID, we used the following tables: observation\_tables, pharma\_records.
Note that events from the emergency department are only available in MIMIC-IV.

The UMCdb and HIRID offer more restricted information compared to MIMIC-IV and eICU, which affects the feasibility of certain tasks or necessitates adjustments. 
For instance, the lack of information in UMCdb for determining the order of ICU stays within a single hospital stay has led us to exclude the readmission task for this dataset. 
Moreover, while the diagnosis codes in MIMIC-IV and eICU were categorized according to the same system used in GenHPF\citep{hur2023genhpf}, those in UMCdb do not align with this categorization, requiring a unique mapping approach.
Regarding HIRID, it does not include ICU discharge times, crucial for filtering cohorts and defining some tasks. 
For cohort filtering, we considered patients with any event recorded more than 48 hours post ICU admission as having met the criteria. 
However, relying on this approximation for task labeling might introduce bias, prompting us to omit mortality and length of stay tasks. 
For reasons akin to those for UMCdb, we excluded the readmission task and applied a different categorization for diagnosis codes.
The diagnosis code categorizations are displayed on Extended Table 4-6.

For both the baseline models and \model, we conducted a grid search for the learning rate, ranging from 1e-6 to 1e-3.
We utilized a constant learning rate scheduler and included 500 warm-up steps.
Early stopping was employed based on the validation AUROC, with patience set to 3 epochs. 
All experiments were performed using an A6000 48G GPU with BF16 mixed precision, and each experiment was repeated using three different random seeds. 
Detailed hyperparameters for the models are provided in \tableref{ext:6}.

\input{tables/hparam}

\section{Recurrent Memory Transformer}\label{apd:rmt}
We also considered using the Recurrent Memory Transformer (RMT) \citep{bulatov2022recurrent}, an architecture that can process virtually unlimited input with constant memory, as the backbone for our baselines.
However, baselines with RMT did not converge unless we adopted a specific training method as the authors suggested \citep{bulatov2023scaling}.
Using this method, which involves learning rate scheduling and curriculum learning, we compared \model to baselines with RMT.
We evaluated those on MIMIC-IV with a 48-hour prediction time setting, which has the longest average input sequence length in our studies.

\begin{figure*}[!htbp]

\begin{minipage}[c]{.5\linewidth}
    \centering
    \includegraphics[clip, width=\textwidth]{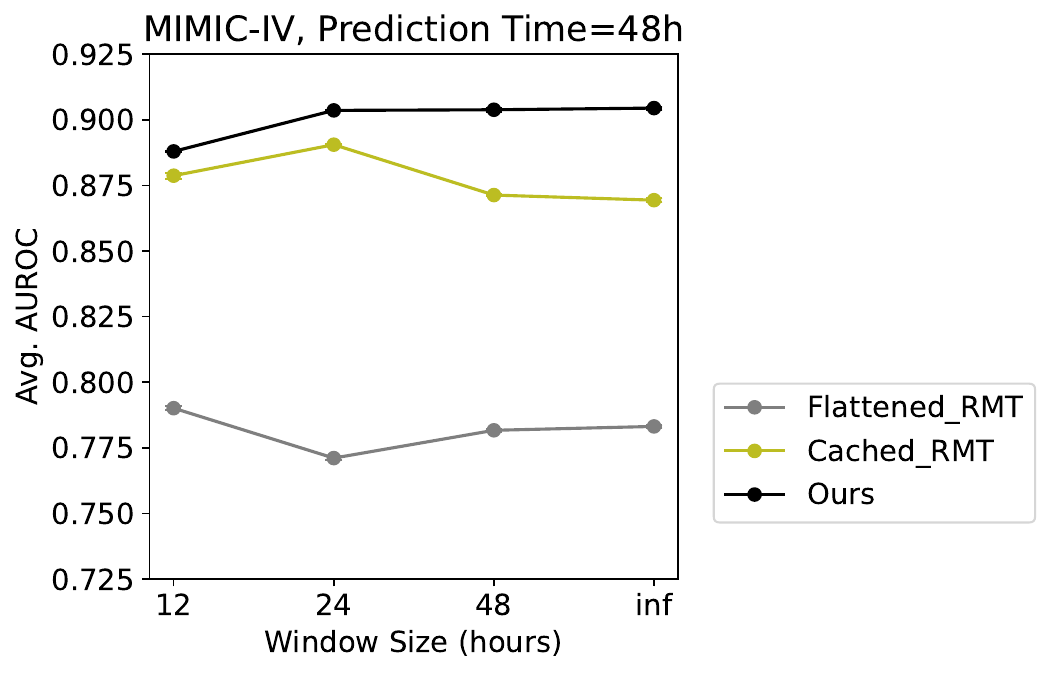}
\end{minipage}%
\begin{minipage}[c]{.5\linewidth}
    \centering
    \includegraphics[clip, width=\textwidth]{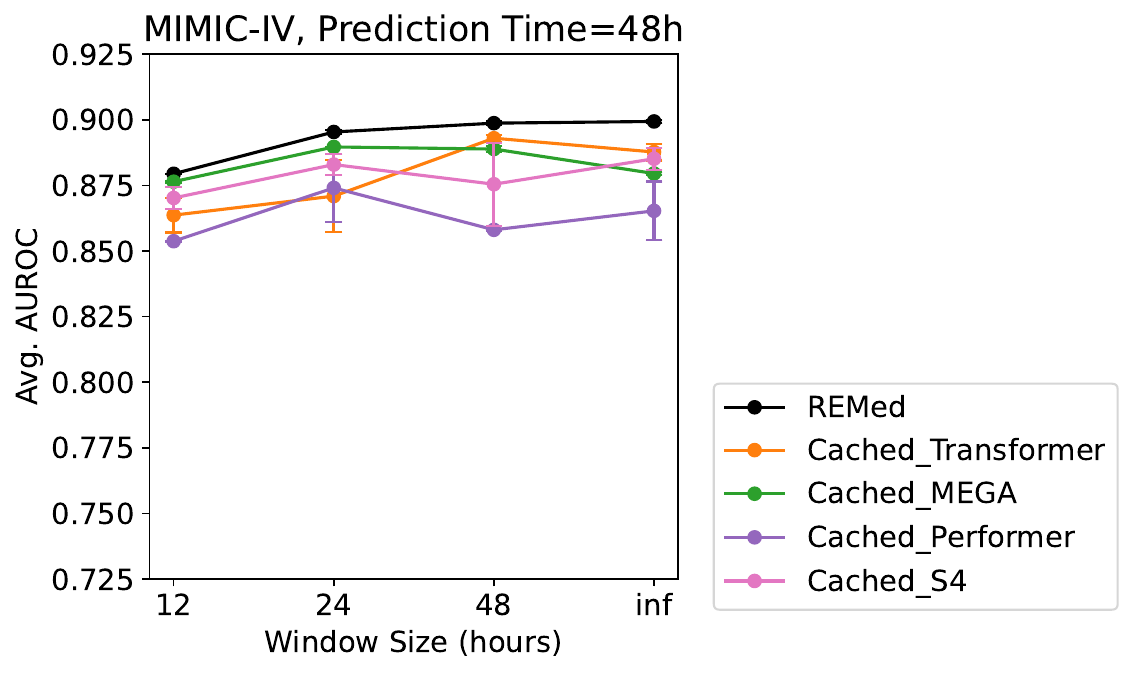}
\end{minipage}
    \caption{(a, left) Comparison with RMT. (b, right) Result for small model size ablation study. The error bars represent the standard error of the mean for three runs with different random seeds.}
    \label{fig:apd}

\end{figure*}

The results are illustrated in Figure \ref{fig:apd} (a). \model's performance remained relatively stable regardless of the training method used, and it consistently surpassed both the \textit{Flattened} and \textit{Cached} RMT (Mann-Whitney U test, $p<0.01$).
Furthermore, as the observation window size expanded, \model showed a monotonic performance increase, even with the addition of curriculum learning and scheduling (Kendall-Tau test, $p<0.01$), while the baseline performances often decreased.

\section{Baselines}\label{apd:baseline}

\textit{GenHPF} \citep{hur2023genhpf}: This approach exploits the inherent hierarchies in EHR data. 
It employs two Transformers: the first one ($\textit{Enc}$) encodes each $r_i$ to a vector $v_i$, while the second one ($\textit{P}$) aggregates these vectors for predictions. 
\begin{equation}\label{eq:hi}
    \hat{y}_{\textit{GenHPF}} = \textit{P}(\{\textit{Enc}(r_i), t_i\})
\end{equation}

\textit{Flattened} model \citep{hur2023genhpf}: This approach chronologically concatenates all $r_i$'s and then feeds them into a Transformer (\textit{P}) for predictions.

\begin{equation}\label{eq:flattened}
 \hat{y}_{\textit{Flattened}} = \textit{P}(\text{Concat}(\{r_i\}), \{t_i\})  
\end{equation}

\textit{Cached} model: This approach utilizes $v_i$'s encoded with the pre-trained text encoder $\textit{Enc}_{\text{PT}}$, similar to that used in \model.
The predictor $\textit{P}$ receives these vectors as input and then makes a prediction. 
The absence of a trainable encoder reduces the computational demands, allowing the model to handle longer sequences.

\begin{equation}\label{eq:cached}
    v_i = \textit{Enc}_{\text{PT}}(r_i),~ \hat{y}_{\textit{Cached}} =\textit{P}(\{v_i, t_i\})    
\end{equation}

To make these models able to handle longer sequences, we used contemporary, efficient architectures \citep{choromanski2020rethinking, gu2021efficiently, ma2022mega} as their backbone (\textit{i.e.}, replacing the vanilla Transformer).
Theoretically, 12 baselines can be derived from these combinations, including the original Transformer version.
However, not all combinations are practical.
For \textit{GenHPF}, the computational bottleneck arises during the event encoding step. 
In this step, the encoder processes numerous $r_i$'s independently, each consisting of several dozen tokens.
Since the efficient architectures do not offer advantages for processing short inputs compared to Transformer, employing them for \textit{GenHPF} is not beneficial.
As a result, we did not replace the Transformer backbone of \textit{GenHPF} with any contemporary architectures.
On the other hand, for the \textit{Flattened} model, using the Transformer backbone is impractical.
The model's strategy—to concatenate all $r_i$'s—yields inputs with at least a few thousand tokens.
Given the quadratic computational complexity of the Transformer, it's infeasible to manage such long inputs using this backbone. 
Therefore, we only used contemporary architectures for the \textit{Flattened} baseline.
In summary, we constructed eight baselines: \textit{GenHPF}-Transformer, \textit{Flattened}-Performer, S4, MEGA, \textit{Cached}-Transformer, Performer, S4, and MEGA.

\section{Additional Performance Analysis}

\subsection{Per-Task Performance Analysis}\label{apd:per_task}
To assess the robustness of REMed across different prediction targets, we evaluated its performance on each task.
We also focused on MIMIC-IV, 48-hour prediction time, and infinite input window scenario.

The results are displayed in \tableref{tab:per_task}. 
Overall, REMed outperformed all baselines in 23 of the 27 tasks, and its performance was comparable to the best baselines in the remaining four tasks.
Based on these results, we conclude that REMed demonstrates strong generalizability across a diverse set of tasks.

\input{tables/per_task}

\subsection{Model Size}
\label{apd:small}

In order to assess \model's robustness with respect to configuration, we expanded our experiment to another model size. 
For simplification, our analysis focused on the MIMIC-IV with a 48-hour prediction time, which has the longest average input length among our test scenarios.
Furthermore, we only considered \textit{Cached} baselines, previously shown to outperform others (\textit{i.e.}, GenHPF and Flattened) in prior experiments.
We configured \model and the baselines with a hidden dimension of 128 and 4 heads, and conducted the same learning rate grid search for each model.
The maximum sequence length for each baseline was adjusted to fit within a 12GB maximum memory allowance.

The results are presented in Figure \ref{fig:apd} (b).
Despite a reduced model size, \model outperformed all baselines in every setting. 
The Mann-Whitney U test \citep{mann1947test} confirmed its superior performance over the best-performing baselines in each setting ($p<0.05$). 
Furthermore, the Kendall-Tau test \citep{kendall1938new} verified a monotonic improvement in \model's performance by increasing the observation window length ($p<0.01$).
These results suggest that \model's key properties hold across different configurations.

\subsection{Comparison to Regression Models}\label{apd:regression}
While machine learning models generally show superior performance compared to regression models, they require more resources for training and evaluation. 
To verify that REMed offers a significant performance benefit over regression models, justifying the complexity, we compared it with additional regression baselines. 
We focused on the MIMIC-IV dataset, with a 48-hour prediction time and an infinite input window scenario.
We constructed logistic, Lasso, Ridge, and Lasso regression models with timestamps. 
The first three models use all unique codes as input features, while the model with timestamps uses bucketized codes categorized by timestamp as input: pre-ICU, 0-24 hours after ICU admission, and 24-48 hours after ICU admission.

As a result, the four models achieved average AUROC of 0.827, 0.841, 0.831, and 0.837, respectively. 
While these performances are comparable to the worst baselines under the same setting in Figure \ref{fig:main_plot}, these scores lag behind REMed's performance (0.903).
Given the need for precise predictions in the clinical domain, REMed's advantages outweigh its complexity.

\section{Extended Top-K Results}\label{apd:topk}
\input{tables/topk_human}

\section{Extended Importance Score Analysis}\label{apd:importance}
\enlargethispage{5\baselineskip}
\rotatebox{90}{
    \begin{minipage}{1\textheight}
        \vspace{8em}
        \includegraphics[clip, width=\textwidth]{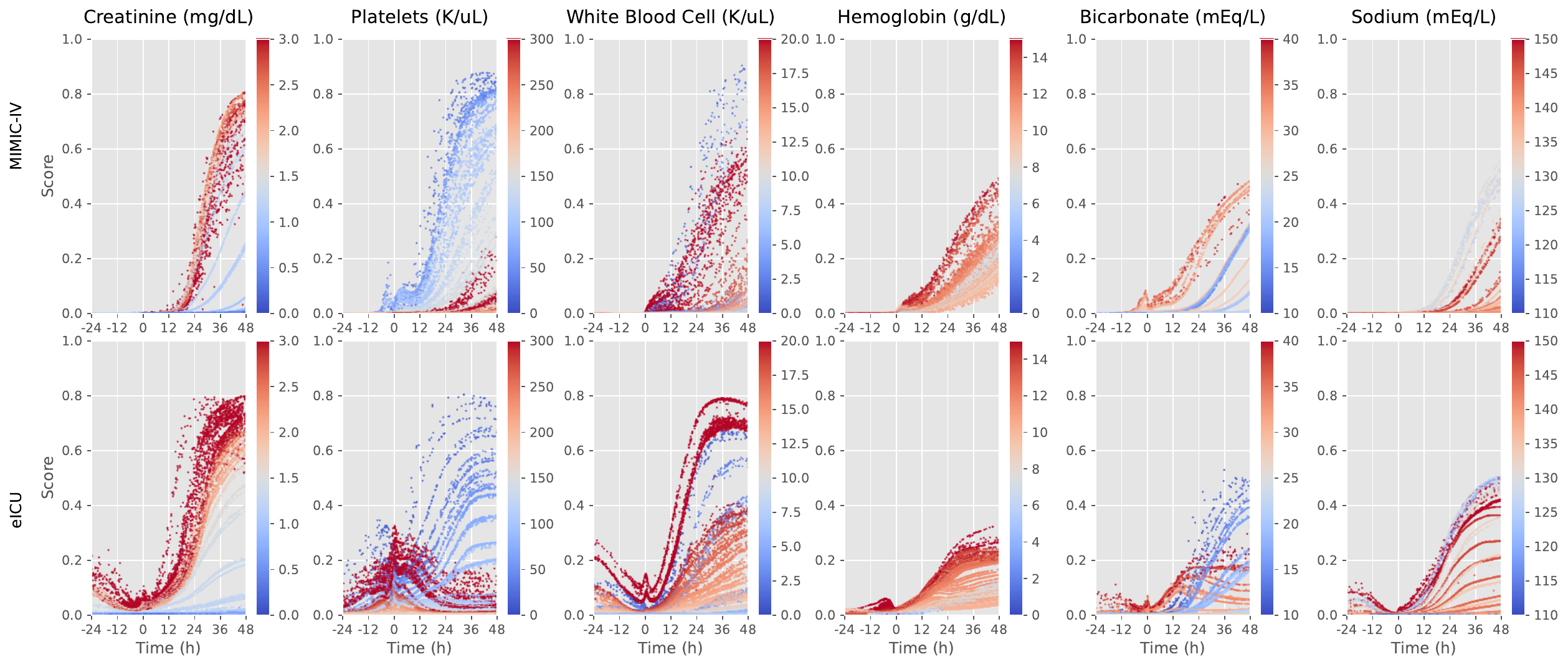}
        \captionof{figure}{Extended Importance Score Analysis Results}
        \label{fig:extended_importance}
    \end{minipage}
}

\end{document}

%% file: tables/task.tex
\begin{table}[!t]
\centering
\footnotesize
\caption{Summary of tasks. For instance, there are two tailored tasks in the Length of Stay category: LOS-7day and LOS-14day.  $^*$, $^{**}$, and $^{***}$ represent binary, multi-class, and multi-label classification tasks, respectively. The additional detail can be found in Section \ref{apd:setting}.}\label{tab:task}
\setlength{\tabcolsep}{0.2em}
\resizebox{\textwidth}{!}{
\begin{tabular}{ll}
\hline
\toprule
\textbf{Category} & \textbf{Description} \\
\hline
\midrule
Mortality & Whether the patient will die within 1/2/3/7/14 days$^*$. \\
\midrule
Length of Stay (LOS) & Whether the length of ICU stay will be longer than 7/14 days$^*$. \\
\midrule
Readmission & Whether the patient will be readmitted to the ICU within the same hospital stay$^*$. \\
\midrule
Diagnosis & Predict all categories of diagnosis codes of the hospital admission$^{**}$. \\
\midrule
Creatinine & Predict the last creatinine measurement value occurred within 1/2/3 days after the prediction time$^{***}$. \\
\midrule
Platelets & Predict the last platelets measurement value occurred within 1/2/3 days after the prediction time$^{***}$. \\
\midrule
White Blood Cells & Predict the last WBC measurement value occurred within 1/2/3 days after the prediction time$^{***}$. \\
\midrule
Hemoglobin & Predict the last hemoglobin measurement value occurred within 1/2/3 days after the prediction time$^{***}$. \\
\midrule
Bicarbonate & Predict the last bicarbonate measurement value occurred within 1/2/3 days after the prediction time$^{***}$. \\
\midrule
Sodium & Predict the last sodium measurement value occurred within 1/2/3 days after the prediction time$^{***}$. \\
\bottomrule
\hline
\end{tabular}
}
\end{table}

%% file: tables/topk.tex
\begin{table*}[!htb]
\caption{The top-30 medical codes most frequently retrieved from MIMIC-IV (left) and eICU (right) are presented. `Avg. Ret.' refers to the average number of times each code appears within the top $k$ retrieved events when REMed makes a prediction. Codes in bold indicate that they are included in the union set of the top-30 codes selected by each clinician. The responses from each expert are displayed in \appendixref{apd:topk}.}\label{tab:topk}
\begin{threeparttable}
\begin{tabular}{@{}ll@{}}
\centering
% \subcaption{MIMIC-IV}
\setlength{\tabcolsep}{0.1em}
\adjustbox{width=0.5\textwidth}{
\begin{tabular}{lrrr}\toprule
\textbf{Table} &\textbf{Code} &\textbf{Avg. Ret} \\\cmidrule{1-3}
labevents &\textbf{Hemoglobin} &4.41 \\\cmidrule{1-3}
labevents &\textbf{Hematocrit} &3.61 \\\cmidrule{1-3}
chartevents &\textbf{Hematocrit (serum)} &3.46 \\\cmidrule{1-3}
labevents &\textbf{Bicarbonate} &2.8 \\\cmidrule{1-3}
labevents &\textbf{Platelet Count} &2.64 \\\cmidrule{1-3}
chartevents &\textbf{HCO$_3$ (serum)} &2.52 \\\cmidrule{1-3}
labevents &\textbf{Creatinine} &2.42 \\\cmidrule{1-3}
chartevents &\textbf{Respiratory Rate} &2.39 \\\cmidrule{1-3}
labevents &\textbf{White Blood Cells} &2.36 \\\cmidrule{1-3}
labevents &\textbf{Sodium} &2.35 \\\cmidrule{1-3}
chartevents &\textbf{WBC}\tnote{1} &2.29 \\\cmidrule{1-3}
chartevents &\textbf{Heart Rhythm} &2.27 \\\cmidrule{1-3}
chartevents &Sodium (serum) &2.26 \\\cmidrule{1-3}
chartevents &\textbf{Creatinine (serum)} &2.15 \\\cmidrule{1-3}
labevents &\textbf{Calculated Total CO$_2$} &2.09 \\\cmidrule{1-3}
chartevents &\textbf{GCS}\tnote{2} \textbf{-Motor Response} &1.95 \\\cmidrule{1-3}
chartevents &Non Invasive BP\tnote{3}  Diastolic &1.9 \\\cmidrule{1-3}
chartevents &\textbf{GCS}\tnote{2} \textbf{-Verbal Response} &1.73 \\\cmidrule{1-3}
chartevents &\textbf{Non Invasive BP}\tnote{3}  \textbf{Systolic} &1.48 \\\cmidrule{1-3}
labevents &Chloride &1.46 \\\cmidrule{1-3}
chartevents &Chloride (serum) &1.45 \\\cmidrule{1-3}
labevents &MCHC\tnote{4} &1.33 \\\cmidrule{1-3}
labevents &Red Blood Cells &1.31 \\\cmidrule{1-3}
chartevents &Pulmonary Artery Pressure Mean &1.1 \\\cmidrule{1-3}
chartevents &\textbf{Mean Airway Pressure} &1.08 \\\cmidrule{1-3}
chartevents &O$_2$ Flow &1.05 \\\cmidrule{1-3}
chartevents &\textbf{Anion Gap} &0.99 \\\cmidrule{1-3}
chartevents &BUN\tnote{5} &0.99 \\\cmidrule{1-3}
outputevents &Foley &0.98 \\\cmidrule{1-3}
chartevents &\textbf{Non Invasive BP}\tnote{3}  \textbf{Mean} &0.96 \\\midrule
\bottomrule
\end{tabular}
}
&

\centering
% \subcaption{eICU}
\setlength{\tabcolsep}{0.1em}
\adjustbox{width=0.48\textwidth}{
\begin{tabular}{lrrr}\toprule
\textbf{Table} &\textbf{Code} &\textbf{Avg. Ret} \\\cmidrule{1-3}
vitalPeriodic &\textbf{vitalPeriodic} &56.87 \\\cmidrule{1-3}
vitalAperiodic &\textbf{vitalAperiodic} &7.99 \\\cmidrule{1-3}
lab &\textbf{Hgb}\tnote{6} &4.88 \\\cmidrule{1-3}
lab &\textbf{Hct}\tnote{7} &4.58 \\\cmidrule{1-3}
lab &\textbf{Creatinine} &4.58 \\\cmidrule{1-3}
lab &\textbf{Platelets$\times$1000} &3.59 \\\cmidrule{1-3}
lab &\textbf{WBC}\tnote{1} \textbf{$\times$1000} &3.46 \\\cmidrule{1-3}
lab &\textbf{Sodium} &3.24 \\\cmidrule{1-3}
lab &RBC\tnote{8} &3.03 \\\cmidrule{1-3}
lab &\textbf{BUN}\tnote{5} &2.53 \\\cmidrule{1-3}
lab &\textbf{Bicarbonate} &2.32 \\\cmidrule{1-3}
lab &Chloride &2.11 \\\cmidrule{1-3}
intakeOutput &Output (ml) $|$ Urine &1.79 \\\cmidrule{1-3}
lab &\textbf{FiO$_2$}\tnote{9} &1.27 \\\cmidrule{1-3}
lab &\textbf{paCO$_2$} &1.16 \\\cmidrule{1-3}
infusionDrug &Propofol (ml/hr) &1.09 \\\cmidrule{1-3}
lab &\textbf{O$_2$ Sat (\%)} &1.00 \\\cmidrule{1-3}
nurseCharting &\textbf{Non-Invasive BP}\tnote{3} &0.90 \\\cmidrule{1-3}
lab &\textbf{HCO$_3$} &0.82 \\\cmidrule{1-3}
lab &\textbf{Base Excess} &0.78 \\\cmidrule{1-3}
infusionDrug &Propofol (mcg/kg/min) &0.71 \\\cmidrule{1-3}
lab &\textbf{Lactate} &0.60 \\\cmidrule{1-3}
intakeOutput &Generic Intake (ml) $|$ NS IVF\tnote{10} &0.59 \\\cmidrule{1-3}
lab &\textbf{paO$_2$} &0.56 \\\cmidrule{1-3}
lab &\textbf{pH} &0.50 \\\cmidrule{1-3}
lab &\textbf{Anion Gap} &0.48 \\\cmidrule{1-3}
infusionDrug &Fentanyl (mcg/hr) &0.48 \\\cmidrule{1-3}
lab &\textbf{Temperature} &0.45 \\\cmidrule{1-3}
infusionDrug &Fentanyl (ml/hr) &0.42 \\\cmidrule{1-3}
lab &PEEP\tnote{11} &0.38 \\\midrule
\bottomrule
\end{tabular}
}
\end{tabular}

\begin{tablenotes}
\setlength{\columnsep}{0.8cm}
\setlength{\multicolsep}{0cm}
  \begin{multicols}{2}
{\footnotesize
\item[1] White Blood Cell
\item[2] Glasgow Coma Scale
\item[3] Blood Pressure
\item[4] Mean Corpuscular Hemoglobin Concentration
\item[5] Blood Urea Nitrogen
\newpage
\item[6] Hemoglobin
\item[7] Hematocrit
\item[8] Red Blood Cell
\item[9] Fraction of Inspired Oxygen
\item[10] Normal Saline Intravenous Fluid
\item[11] Positive End-Expiratory Pressure
}
  \end{multicols}

\end{tablenotes}
\end{threeparttable}
\vspace{-1em}

\end{table*}

%% file: tables/length.tex
\begin{table*}[!htp]\centering
\caption{Data Input Length Distribution}\label{ext:3}
\scriptsize
\begin{adjustbox}{width=1.0\textwidth}
\setlength{\tabcolsep}{0.2em}
\begin{tabular}{lrrrrrrrrrrr}\toprule
\multirow{2}{*}{Dataset} &\multirow{2}{*}{Prediction Time (h)} &\multirow{2}{*}{Input Horizon (h)} &\multicolumn{4}{c}{\# of Events} &\multicolumn{4}{c}{\# of Tokens} \\\cmidrule{4-11}
& & &average &median &90\% &max &average &median &90\% &max \\\cmidrule{1-11}
\multirow{8}{*}{MIMIC-IV} &\multirow{4}{*}{24} &6 &344.9 &327 &539 &1832 &8002.9 &7266 &13149 &48368 \\\cmidrule{3-11}
& &12 &703.2 &666 &1056 &3146 &16534.0 &15054 &26381.4 &83492 \\\cmidrule{3-11}
& &24 &1601.2 &1545 &2279 &5457 &39454.1 &36599 &60227 &148723 \\\cmidrule{3-11}
& &inf &2713.6 &1748 &3289.4 &264838 &67697.2 &43441 &90027.2 &6621844 \\\cmidrule{2-11}
&\multirow{4}{*}{48} &12 &621.7 &578 &948 &3198 &13988.7 &12539 &22314 &83327 \\\cmidrule{3-11}
& &24 &1278.7 &1188 &1907 &6409 &29017.6 &26035 &45289.6 &167210 \\\cmidrule{3-11}
& &48 &2879.9 &2772 &4081 &10928 &68471.7 &63959 &101734.4 &286674 \\\cmidrule{3-11}
& &inf &3992.4 &3023 &5106.2 &266766 &96714.8 &72443 &133924 &6669451 \\\cmidrule{1-11}
\multirow{8}{*}{eICU} &\multirow{4}{*}{24} &6 &175.7 &153 &293 &1135 &6766.3 &5695 &12358 &43989 \\\cmidrule{3-11}
& &12 &360.3 &315 &593 &2225 &13931.8 &11831 &25173.3 &83519 \\\cmidrule{3-11}
& &24 &774.8 &687 &1246 &4825 &30026.3 &25898.5 &53245 &178635 \\\cmidrule{3-11}
& &inf &1290.7 &824 &1838 &140592 &50954.9 &32083 &77627.6 &5640899 \\\cmidrule{2-11}
&\multirow{4}{*}{48} &12 &329.7 &293 &544 &4039 &12532.3 &10662 &22923.3 &166365 \\\cmidrule{3-11}
& &24 &673.0 &600 &1096 &4577 &25715.8 &22068 &46326.2 &187296 \\\cmidrule{3-11}
& &48 &1447.8 &1303 &2318 &8159 &55742.1 &48499.5 &97743 &304656 \\\cmidrule{3-11}
& &inf &1963.7 &1464.5 &3021 &141509 &76670.6 &55958 &126776.3 &5673689 \\\cmidrule{1-11}
\multirow{4}{*}{UMCdb} &\multirow{4}{*}{48} &12 &6293.7 &619 &23887.6 &57080 &199068.8 &31736 &717089.8 &1781734 \\\cmidrule{3-11}
& &24 &2816.2 &1241 &48189.2 &113830 &404977.9 &63641 &1456686.4 &3552063 \\\cmidrule{3-11}
& &48 &26443.7 &2559 &95681.8 &212686 &835479.8 &131369 &2915309.2 &6671146 \\\cmidrule{3-11}
& &inf &26591.4 &2698 &96039.4 &212766 &842346.4 &137966 &2933463.6 &6674956 \\\cmidrule{1-11}
\multirow{4}{*}{HIRID} &\multirow{4}{*}{48} &12 &5315.1 &5350 &9212.2 &15811 &109802.0 &108561 &192767 &321138 \\\cmidrule{3-11}
& &24 &10941.5 &11179 &18422 &31386 &225919.5 &226896 &385757.8 &632670 \\\cmidrule{3-11}
& &48 &22328.0 &22620 &35702.2 &60858 &460126.2 &461563 &746492 &1235864 \\\cmidrule{3-11}
& &inf &22332.8 &22629 &35703.8 &60858 &460233.6 &461572 &746519.2 &1235864 \\\midrule
\bottomrule
\end{tabular}
\end{adjustbox}
\end{table*}

%% file: tables/stat.tex
\begin{table*}[!htbp]\centering
\caption{MIMIC-IV and eICU Data Statistics and Diagnosis Label Distribution}
\footnotesize
\setlength{\tabcolsep}{0.2em}

\begin{tabular}{lrrrrrrrrrrrrrrrrrrrrrrrrrr}\toprule
\multicolumn{2}{c}{Datasets} &\multicolumn{12}{c}{MIMIC-IV} &\multicolumn{12}{c}{eICU} \\\cmidrule{1-26}
\multicolumn{2}{c}{Number of patients} &\multicolumn{12}{c}{25801} &\multicolumn{12}{c}{64276} \\\cmidrule{1-26}
\multicolumn{2}{c}{Number of admissions} &\multicolumn{12}{c}{30360} &\multicolumn{12}{c}{72298} \\\cmidrule{1-26}
\multicolumn{2}{c}{Number of icustays} &\multicolumn{12}{c}{32449} &\multicolumn{12}{c}{77718} \\\cmidrule{1-26}
\multicolumn{2}{c}{Avg admission duration (d)} &\multicolumn{12}{c}{11.8} &\multicolumn{12}{c}{9.3} \\\cmidrule{1-26}
\multicolumn{2}{c}{Avg ICU stay duration (d)} &\multicolumn{12}{c}{5.6} &\multicolumn{12}{c}{5.4} \\\cmidrule{1-26}
\multicolumn{2}{c}{Avg. Age} &\multicolumn{12}{c}{64.1} &\multicolumn{12}{c}{64.3} \\\cmidrule{1-26}
\multirow{3}{*}{Gender} &Male &\multicolumn{12}{c}{18280} &\multicolumn{12}{c}{42460} \\\cmidrule{2-26}
&Female &\multicolumn{12}{c}{14169} &\multicolumn{12}{c}{35244} \\\cmidrule{2-26}
&Others/Unknown &\multicolumn{12}{c}{-} &\multicolumn{12}{c}{7} \\\cmidrule{1-26}
\multirow{5}{*}{Ethinicity} &White &\multicolumn{12}{c}{21981} &\multicolumn{12}{c}{59706} \\\cmidrule{2-26}
&Black &\multicolumn{12}{c}{3318} &\multicolumn{12}{c}{9146} \\\cmidrule{2-26}
&Hispanic &\multicolumn{12}{c}{1163} &\multicolumn{12}{c}{2866} \\\cmidrule{2-26}
&Asian &\multicolumn{12}{c}{930} &\multicolumn{12}{c}{1290} \\\cmidrule{2-26}
&Others/Unknown &\multicolumn{12}{c}{5057} &\multicolumn{12}{c}{3957} \\\cmidrule{1-26}
\midrule
\multicolumn{26}{c}{Diagnosis Label Distrubution (Multilabel)} \\\cmidrule{1-26}
\midrule
\multicolumn{2}{c}{Infectious and parasitic diseases} &\multicolumn{12}{c}{14848} &\multicolumn{12}{c}{13899} \\\cmidrule{1-26}
\multicolumn{2}{c}{Neoplasms} &\multicolumn{12}{c}{10607} &\multicolumn{12}{c}{4979} \\\cmidrule{1-26}
\multicolumn{2}{c}{\makecell{Endocrine; nutritional; and metabolic diseases and immunity disorders}} &\multicolumn{12}{c}{27826} &\multicolumn{12}{c}{22248} \\\cmidrule{1-26}
\multicolumn{2}{c}{Diseases of the blood and blood-forming organs} &\multicolumn{12}{c}{19885} &\multicolumn{12}{c}{11958} \\\cmidrule{1-26}
\multicolumn{2}{c}{Mental Illness} &\multicolumn{12}{c}{20370} &\multicolumn{12}{c}{10303} \\\cmidrule{1-26}
\multicolumn{2}{c}{Diseases of the nervous system and sense organs} &\multicolumn{12}{c}{17189} &\multicolumn{12}{c}{14842} \\\cmidrule{1-26}
\multicolumn{2}{c}{Diseases of the circulatory system} &\multicolumn{12}{c}{29491} &\multicolumn{12}{c}{45252} \\\cmidrule{1-26}
\multicolumn{2}{c}{Diseases of the respiratory system} &\multicolumn{12}{c}{21338} &\multicolumn{12}{c}{35259} \\\cmidrule{1-26}
\multicolumn{2}{c}{Diseases of the digestive system} &\multicolumn{12}{c}{20129} &\multicolumn{12}{c}{13803} \\\cmidrule{1-26}
\multicolumn{2}{c}{Diseases of the genitourinary system} &\multicolumn{12}{c}{20411} &\multicolumn{12}{c}{21039} \\\cmidrule{1-26}
\multicolumn{2}{c}{\makecell{Complications of pregnancy;\\ childbirth; and the puerperium}} &\multicolumn{12}{c}{141} &\multicolumn{12}{c}{141} \\\cmidrule{1-26}
\multicolumn{2}{c}{Diseases of the skin and subcutaneous tissue} &\multicolumn{12}{c}{4856} &\multicolumn{12}{c}{1617} \\\cmidrule{1-26}
\multicolumn{2}{c}{\makecell{Diseases of the musculoskeletal system and connective tissue}} &\multicolumn{12}{c}{11483} &\multicolumn{12}{c}{1835} \\\cmidrule{1-26}
\multicolumn{2}{c}{Congenital anomalies} &\multicolumn{12}{c}{1504} &\multicolumn{12}{c}{59} \\\cmidrule{1-26}
\multicolumn{2}{c}{Injury and poisoning} &\multicolumn{12}{c}{16867} &\multicolumn{12}{c}{14150} \\\cmidrule{1-26}
\multicolumn{2}{c}{\makecell{Symptoms; signs; and ill-defined conditions and factors influencing health status}} &\multicolumn{12}{c}{18741} &\multicolumn{12}{c}{13179} \\\cmidrule{1-26}
\multicolumn{2}{c}{\makecell{Residual codes; unclassified; all E codes [259. and 260.]}} &\multicolumn{12}{c}{24490} &\multicolumn{12}{c}{6665} \\\cmidrule{1-26}
\bottomrule
\end{tabular}
\end{table*}

\begin{table*}[!htbp]\centering
\caption{UMCdb Data Statistics and Diagnosis Label Distribution}
\footnotesize
\setlength{\tabcolsep}{0.2em}

\begin{tabular}{lrrrrrrrr}\toprule
\multicolumn{2}{c}{Datasets} &\multicolumn{6}{c}{UMCdb} \\\cmidrule{1-8}
\multicolumn{2}{c}{Number of patients} &\multicolumn{6}{c}{7392} \\\cmidrule{1-8}
\multicolumn{2}{c}{Number of admissions} &\multicolumn{6}{c}{-} \\\cmidrule{1-8}
\multicolumn{2}{c}{Number of icustays} &\multicolumn{6}{c}{8359} \\\cmidrule{1-8}
Avg length of admissions (d) & &\multicolumn{6}{c}{-} \\\cmidrule{1-1}\cmidrule{3-8}
Avg length of icustays & &\multicolumn{6}{c}{11.1187941141284} \\\cmidrule{1-1}\cmidrule{3-8}
\multicolumn{2}{c}{Avg. Age} &\multicolumn{6}{c}{-} \\\cmidrule{1-8}
\multirow{3}{*}{Gender} &Male &\multicolumn{6}{c}{5223} \\\cmidrule{2-8}
&Female &\multicolumn{6}{c}{2986} \\\cmidrule{2-8}
&Others/Unknown &\multicolumn{6}{c}{-} \\\cmidrule{1-8}
\multirow{5}{*}{Ethinicity} &White &\multicolumn{6}{c}{-} \\\cmidrule{2-8}
&Black &\multicolumn{6}{c}{-} \\\cmidrule{2-8}
&Hispanic &\multicolumn{6}{c}{-} \\\cmidrule{2-8}
&Asian &\multicolumn{6}{c}{-} \\\cmidrule{2-8}
&Others/Unknown &\multicolumn{6}{c}{-} \\\cmidrule{1-8}
\multicolumn{8}{c}{Diagnosis Label Distrubution (Multilabel)} \\\cmidrule{1-8}
\multicolumn{2}{c}{Cardiovascular} &\multicolumn{6}{c}{1733} \\\cmidrule{1-8}
\multicolumn{2}{c}{General Surgery} &\multicolumn{6}{c}{3083} \\\cmidrule{1-8}
\multicolumn{2}{c}{Respiratory} &\multicolumn{6}{c}{856} \\\cmidrule{1-8}
\multicolumn{2}{c}{Neurological} &\multicolumn{6}{c}{1096} \\\cmidrule{1-8}
\multicolumn{2}{c}{Genitourinary/Renal} &\multicolumn{6}{c}{124} \\\cmidrule{1-8}
\multicolumn{2}{c}{Gastrointestinal} &\multicolumn{6}{c}{444} \\\cmidrule{1-8}
\multicolumn{2}{c}{Hematological} &\multicolumn{6}{c}{73} \\\cmidrule{1-8}
\multicolumn{2}{c}{Transplant} &\multicolumn{6}{c}{18} \\\cmidrule{1-8}
\multicolumn{2}{c}{Trauma} &\multicolumn{6}{c}{376} \\\cmidrule{1-8}
\multicolumn{2}{c}{Metabolic} &\multicolumn{6}{c}{88} \\\cmidrule{1-8}
\multicolumn{2}{c}{Musculoskeletal/Skin} &\multicolumn{6}{c}{56} \\\cmidrule{1-8}
\multicolumn{2}{c}{Internal Medicine} &\multicolumn{6}{c}{2770} \\\cmidrule{1-8}
\multicolumn{2}{c}{Non-Categorized/General} &\multicolumn{6}{c}{3491} \\\midrule
\bottomrule
\end{tabular}
\end{table*}

\begin{table*}[!htbp]\centering
\caption{HIRID Data Statistics and Diagnosis Label Distribution}
\footnotesize
\setlength{\tabcolsep}{0.2em}

\begin{tabular}{lrrrrrrrr}\toprule
\multicolumn{2}{c}{Datasets} &\multicolumn{6}{c}{HIRID} \\\cmidrule{1-8}
\multicolumn{2}{c}{Number of patients} &\multicolumn{6}{c}{-} \\\cmidrule{1-8}
\multicolumn{2}{c}{Number of admissions} &\multicolumn{6}{c}{-} \\\cmidrule{1-8}
\multicolumn{2}{c}{Number of icustays} &\multicolumn{6}{c}{9155} \\\cmidrule{1-8}
Avg length of admissions (d) & &\multicolumn{6}{c}{-} \\\cmidrule{1-1}\cmidrule{3-8}
Avg length of icustays & &\multicolumn{6}{c}{-} \\\cmidrule{1-1}\cmidrule{3-8}
\multicolumn{2}{c}{Avg. Age} &\multicolumn{6}{c}{62.2725286728563} \\\cmidrule{1-8}
\multirow{3}{*}{Gender} &Male &\multicolumn{6}{c}{5848} \\\cmidrule{2-8}
&Female &\multicolumn{6}{c}{3307} \\\cmidrule{2-8}
&Others/Unknown &\multicolumn{6}{c}{-} \\\cmidrule{1-8}
\multirow{5}{*}{Ethinicity} &White &\multicolumn{6}{c}{-} \\\cmidrule{2-8}
&Black &\multicolumn{6}{c}{-} \\\cmidrule{2-8}
&Hispanic &\multicolumn{6}{c}{-} \\\cmidrule{2-8}
&Asian &\multicolumn{6}{c}{-} \\\cmidrule{2-8}
&Others/Unknown &\multicolumn{6}{c}{-} \\\cmidrule{1-8}
\multicolumn{8}{c}{Diagnosis Label Distrubution (Multilabel)} \\\cmidrule{1-8}
\multicolumn{2}{c}{Cardiovascular} &\multicolumn{6}{c}{1893} \\\cmidrule{1-8}
\multicolumn{2}{c}{Pulmonary} &\multicolumn{6}{c}{1178} \\\cmidrule{1-8}
\multicolumn{2}{c}{Gastointestinal} &\multicolumn{6}{c}{1048} \\\cmidrule{1-8}
\multicolumn{2}{c}{Neurological} &\multicolumn{6}{c}{2410} \\\cmidrule{1-8}
\multicolumn{2}{c}{Sepsis + Intoxication} &\multicolumn{6}{c}{354} \\\cmidrule{1-8}
\multicolumn{2}{c}{Urogenital} &\multicolumn{6}{c}{24} \\\cmidrule{1-8}
\multicolumn{2}{c}{Trauma} &\multicolumn{6}{c}{806} \\\cmidrule{1-8}
\multicolumn{2}{c}{Metabolic/Endocrinology} &\multicolumn{6}{c}{198} \\\cmidrule{1-8}
\multicolumn{2}{c}{Hematology} &\multicolumn{6}{c}{51} \\\cmidrule{1-8}
\multicolumn{2}{c}{Other} &\multicolumn{6}{c}{186} \\\cmidrule{1-8}
\multicolumn{2}{c}{Surgical Cardiovascular} &\multicolumn{6}{c}{1010} \\\cmidrule{1-8}
\multicolumn{2}{c}{Surgical Respiratory} &\multicolumn{6}{c}{375} \\\cmidrule{1-8}
\multicolumn{2}{c}{Surgical Gastrointestinal} &\multicolumn{6}{c}{256} \\\cmidrule{1-8}
\multicolumn{2}{c}{Surgical Neurological} &\multicolumn{6}{c}{896} \\\cmidrule{1-8}
\multicolumn{2}{c}{Surgical Trauma} &\multicolumn{6}{c}{264} \\\cmidrule{1-8}
\multicolumn{2}{c}{Surgical Urogenital} &\multicolumn{6}{c}{17} \\\cmidrule{1-8}
\multicolumn{2}{c}{Surgical Others} &\multicolumn{6}{c}{148} \\\midrule
\bottomrule
\end{tabular}
\end{table*}

%% file: tables/label.tex
\renewcommand{\arraystretch}{0.5}

\begin{table*}
\caption{MIMIC-IV and eICU Label Distribution (binary/multiclass)}
\begin{adjustbox}{width=0.9\textwidth, center}
%\scriptsize
\setlength{\tabcolsep}{0.2em}
\begin{tabular}{lrrrrrrrrrrrrrrr}\toprule
\multicolumn{3}{c}{Datasets} &\multicolumn{6}{c}{MIMIC-IV} &\multicolumn{6}{c}{eICU} \\\cmidrule{1-15}
\makecell{Prediction\\Time (h)} &Task &\makecell{Prediction\\Window} &0 &1 &2 &3 &4 &NaN &0 &1 &2 &3 &4 &NaN \\\cmidrule{1-15}
\multirow{26}{*}{24} &Readmission &- &30360 &2089 &0 &0 &0 &0 &72298 &5420 &0 &0 &0 &0 \\\cmidrule{2-15}
&\multirow{2}{*}{Length of Stay} &7 &25887 &6562 &0 &0 &0 &0 &62765 &14953 &0 &0 &0 &0 \\\cmidrule{3-15}
& &14 &30374 &2075 &0 &0 &0 &0 &73508 &4210 &0 &0 &0 &0 \\\cmidrule{2-15}
&\multirow{5}{*}{Mortality} &1 &32296 &153 &0 &0 &0 &0 &77702 &16 &0 &0 &0 &0 \\\cmidrule{3-15}
& &2 &31730 &719 &0 &0 &0 &0 &76452 &1266 &0 &0 &0 &0 \\\cmidrule{3-15}
& &3 &31283 &1166 &0 &0 &0 &0 &75341 &2377 &0 &0 &0 &0 \\\cmidrule{3-15}
& &7 &30010 &2439 &0 &0 &0 &0 &72553 &5165 &0 &0 &0 &0 \\\cmidrule{3-15}
& &14 &28935 &3514 &0 &0 &0 &0 &70414 &7304 &0 &0 &0 &0 \\\cmidrule{2-15}
&\multirow{3}{*}{Creatinine} &1 &18799 &6234 &2784 &889 &611 &3132 &41597 &13775 &6993 &2389 &2030 &10934 \\\cmidrule{3-15}
& &2 &18461 &5471 &2297 &800 &496 &4924 &39847 &11836 &5903 &2049 &1768 &16315 \\\cmidrule{3-15}
& &3 &17150 &4895 &1945 &660 &396 &7403 &34624 &9760 &4692 &1685 &1501 &25456 \\\cmidrule{2-15}
&\multirow{3}{*}{Platelets} &1 &18751 &7277 &4054 &1077 &167 &1123 &41873 &15053 &8269 &1929 &362 &10232 \\\cmidrule{3-15}
& &2 &17927 &6680 &3889 &1110 &172 &2671 &37890 &13862 &7831 &1968 &415 &15752 \\\cmidrule{3-15}
& &3 &17870 &5204 &2999 &1012 &174 &5190 &33430 &10651 &6228 &1734 &381 &25294 \\\cmidrule{2-15}
&\multirow{3}{*}{WBC} &1 &1191 &20320 &9805 &0 &0 &1133 &2182 &42804 &22931 &0 &0 &9801 \\\cmidrule{3-15}
& &2 &1275 &21064 &7439 &0 &0 &2671 &2026 &42260 &18051 &0 &0 &15381 \\\cmidrule{3-15}
& &3 &1335 &20219 &5722 &0 &0 &5173 &1794 &37480 &13476 &0 &0 &24968 \\\cmidrule{2-15}
&\multirow{3}{*}{Hemoglobin} &1 &4079 &13071 &9335 &4838 &0 &1126 &7945 &25816 &20906 &14003 &0 &9048 \\\cmidrule{3-15}
& &2 &3956 &13048 &8620 &4153 &0 &2672 &7413 &25365 &18945 &11316 &0 &14679 \\\cmidrule{3-15}
& &3 &3597 &12223 &7875 &3560 &0 &5194 &5852 &22546 &16088 &8920 &0 &24312 \\\cmidrule{2-15}
&\multirow{3}{*}{Bicarbonate} &1 &7933 &18764 &4788 &0 &0 &964 &15165 &38746 &12695 &0 &0 &11112 \\\cmidrule{3-15}
& &2 &6069 &17624 &6307 &0 &0 &2449 &11439 &35642 &14353 &0 &0 &16284 \\\cmidrule{3-15}
& &3 &4519 &15809 &7067 &0 &0 &5054 &7926 &29860 &14387 &0 &0 &25545 \\\cmidrule{2-15}
&\multirow{3}{*}{Sodium} &1 &5986 &22695 &2855 &0 &0 &913 &11651 &50147 &8744 &0 &0 &7176 \\\cmidrule{3-15}
& &2 &5313 &21790 &2967 &0 &0 &2379 &10415 &45807 &8964 &0 &0 &12532 \\\cmidrule{3-15}
& &3 &4297 &20228 &3114 &0 &0 &4810 &8458 &38904 &8236 &0 &0 &22120 \\\cmidrule{1-15}
\multirow{26}{*}{48} &Readmission &- &30360 &2089 &0 &0 &0 &0 &72298 &5420 &0 &0 &0 &0 \\\cmidrule{2-15}
&\multirow{2}{*}{Length of Stay} &7 &25887 &6562 &0 &0 &0 &0 &62765 &14953 &0 &0 &0 &0 \\\cmidrule{3-15}
& &14 &30374 &2075 &0 &0 &0 &0 &73508 &4210 &0 &0 &0 &0 \\\cmidrule{2-15}
&\multirow{5}{*}{Mortality} &1 &31730 &719 &0 &0 &0 &0 &76452 &1266 &0 &0 &0 &0 \\\cmidrule{3-15}
& &2 &31283 &1166 &0 &0 &0 &0 &75341 &2377 &0 &0 &0 &0 \\\cmidrule{3-15}
& &3 &30884 &1565 &0 &0 &0 &0 &74413 &3305 &0 &0 &0 &0 \\\cmidrule{3-15}
& &7 &29803 &2646 &0 &0 &0 &0 &72116 &5602 &0 &0 &0 &0 \\\cmidrule{3-15}
& &14 &28834 &3615 &0 &0 &0 &0 &70249 &7469 &0 &0 &0 &0 \\\cmidrule{2-15}
&\multirow{3}{*}{Creatinine} &1 &18461 &5471 &2297 &800 &496 &4924 &39847 &11836 &5903 &2049 &1768 &16315 \\\cmidrule{3-15}
& &2 &17150 &4895 &1945 &660 &396 &7403 &34624 &9760 &4692 &1685 &1501 &25456 \\\cmidrule{3-15}
& &3 &15053 &4268 &1676 &516 &331 &10605 &29176 &8113 &3941 &1386 &1256 &33846 \\\cmidrule{2-15}
&\multirow{3}{*}{Platelets} &1 &17927 &6680 &3889 &1110 &172 &2671 &33430 &10651 &6228 &1734 &381 &25294 \\\cmidrule{3-15}
& &2 &17870 &5204 &2999 &1012 &174 &5190 &29359 &7816 &4714 &1501 &337 &33991 \\\cmidrule{3-15}
& &3 &16579 &3843 &2394 &898 &153 &8582 &2026 &42260 &18051 &0 &0 &15381 \\\cmidrule{2-15}
&\multirow{3}{*}{WBC} &1 &1275 &21064 &7439 &0 &0 &2671 &1794 &37480 &13476 &0 &0 &24968 \\\cmidrule{3-15}
& &2 &1335 &20219 &5722 &0 &0 &5173 &1500 &31226 &11301 &0 &0 &33691 \\\cmidrule{3-15}
& &3 &1218 &17782 &4895 &0 &0 &8554 &7413 &25365 &18945 &11316 &0 &14679 \\\cmidrule{2-15}
&\multirow{3}{*}{Hemoglobin} &1 &3956 &13048 &8620 &4153 &0 &2672 &5852 &22546 &16088 &8920 &0 &24312 \\\cmidrule{3-15}
& &2 &3597 &12223 &7875 &3560 &0 &5194 &4840 &19266 &13302 &7150 &0 &33160 \\\cmidrule{3-15}
& &3 &3215 &10855 &6795 &3007 &0 &8577 &11439 &35642 &14353 &0 &0 &16284 \\\cmidrule{2-15}
&\multirow{3}{*}{Bicarbonate} &1 &6069 &17624 &6307 &0 &0 &2449 &7926 &29860 &14387 &0 &0 &25545 \\\cmidrule{3-15}
& &2 &4519 &15809 &7067 &0 &0 &5054 &5767 &24342 &13752 &0 &0 &33857 \\\cmidrule{3-15}
& &3 &3528 &13634 &6878 &0 &0 &8409 &10415 &45807 &8964 &0 &0 &12532 \\\cmidrule{2-15}
&\multirow{3}{*}{Sodium} &1 &5313 &21790 &2967 &0 &0 &2379 &8458 &38904 &8236 &0 &0 &22120 \\\cmidrule{3-15}
& &2 &4297 &20228 &3114 &0 &0 &4810 &6936 &32619 &7337 &0 &0 &30826 \\\cmidrule{3-15}
& &3 &3656 &17709 &2981 &0 &0 &8103 &8458 &38904 &8236 &0 &0 &22120 \\\cmidrule{1-15}
\bottomrule
\end{tabular}
\end{adjustbox}
\end{table*}

\renewcommand{\arraystretch}{0.5}

\begin{table*}
\caption{UMCdb and HIRID Label Distribution (binary/multiclass)}
\begin{adjustbox}{width=1.0\textwidth, center}
%\scriptsize
\setlength{\tabcolsep}{0.2em}
\begin{tabular}{lrrrrrrrrrrrrrr}\toprule
\multicolumn{2}{c}{Datasets} &\multicolumn{6}{c}{UMCdb} &\multicolumn{6}{c}{HIRID} \\\cmidrule{1-14}
Task &Prediction Window &0 &1 &2 &3 &4 &NaN &0 &1 &2 &3 &4 &NaN \\\cmidrule{1-14}
Readmission &- &- &- &- &- &- &- &- &- &- &- &- &- \\\cmidrule{1-14}
\multirow{2}{*}{Length of Stay} &7 &4705 &3654 &0 &0 &0 &0 &- &- &- &- &- &- \\\cmidrule{2-14}
&14 &6411 &1948 &0 &0 &0 &0 &- &- &- &- &- &- \\\cmidrule{1-14}
\multirow{5}{*}{Mortality} &1 &8086 &273 &0 &0 &0 &0 &- &- &- &- &- &- \\\cmidrule{2-14}
&2 &7896 &463 &0 &0 &0 &0 &- &- &- &- &- &- \\\cmidrule{2-14}
&3 &7785 &574 &0 &0 &0 &0 &- &- &- &- &- &- \\\cmidrule{2-14}
&7 &7510 &849 &0 &0 &0 &0 &- &- &- &- &- &- \\\cmidrule{2-14}
&14 &7251 &1108 &0 &0 &0 &0 &- &- &- &- &- &- \\\cmidrule{1-14}
\multirow{3}{*}{Creatinine} &1 &4007 &714 &348 &88 &30 &3172 &4321 &912 &516 &150 &49 &3207 \\\cmidrule{2-14}
&2 &3056 &500 &231 &60 &23 &4489 &2989 &590 &278 &80 &36 &5182 \\\cmidrule{2-14}
&3 &2495 &401 &157 &42 &19 &5245 &2159 &409 &152 &39 &21 &6375 \\\cmidrule{1-14}
\multirow{3}{*}{Platelets} &1 &4080 &1679 &1138 &360 &110 &992 &3647 &1545 &1310 &426 &141 &2086 \\\cmidrule{2-14}
&2 &3368 &1218 &861 &318 &105 &2489 &2711 &983 &789 &352 &121 &4199 \\\cmidrule{2-14}
&3 &3071 &870 &648 &290 &87 &3393 &2219 &612 &471 &276 &91 &5486 \\\cmidrule{1-14}
\multirow{3}{*}{WBC} &1 &208 &1108 &3178 &0 &0 &3865 &254 &4982 &1925 &0 &0 &1994 \\\cmidrule{2-14}
&2 &168 &3333 &2368 &0 &0 &2490 &176 &3591 &1252 &0 &0 &4136 \\\cmidrule{2-14}
&3 &135 &2889 &1943 &0 &0 &3392 &134 &2582 &1026 &0 &0 &5413 \\\cmidrule{1-14}
\multirow{3}{*}{Hemoglobin} &1 &418 &3284 &2791 &1145 &0 &721 &509 &3622 &2155 &864 &0 &2005 \\\cmidrule{2-14}
&2 &313 &2609 &2255 &847 &0 &2335 &306 &2654 &1513 &545 &0 &4137 \\\cmidrule{2-14}
&3 &272 &2198 &1903 &713 &0 &3273 &222 &2005 &1121 &389 &0 &5418 \\\cmidrule{1-14}
\multirow{3}{*}{Bicarbonate} &1 &1661 &4486 &1271 &0 &0 &941 &823 &4820 &983 &0 &0 &2529 \\\cmidrule{2-14}
&2 &921 &3502 &1461 &0 &0 &2475 &426 &3285 &882 &0 &0 &4562 \\\cmidrule{2-14}
&3 &543 &2880 &1557 &0 &0 &3379 &303 &2357 &734 &0 &0 &5761 \\\midrule
\bottomrule
\end{tabular}
\end{adjustbox}
\end{table*}

%% file: tables/hparam.tex
\begin{table*}[!ht]\centering
\caption{Model Hyperparameters. For all of the models, we used batch size 8, 512 hidden dimensions.
REMed is capable of processing a practically infinite number of events, even can handle the longest case in our dataset. Since both the REMed and the Cached models use a 2-layer pre-trained event encoder, we ensured that the total number of layers was matched.
}\label{ext:6}
\scriptsize
\begin{adjustbox}{width=\textwidth, center}
\begin{tabular}{lrrrrrrr}\toprule
\multicolumn{2}{c}{Models} &Max. Tokens &Max. Events &LR &\makecell{No. of Layers\\(Trainable)} &Model Size (M) \\\cmidrule{1-7}
\multirow{3}{*}{Flatten} &Mega &8192 &- &5e-4 &4 &28.1 \\\cmidrule{2-7}
&S4 &16384 &- &1e-4 &4 &28.9 \\\cmidrule{2-7}
&Performer &16384 &- &5e-5 &4 &27.6 \\\cmidrule{1-7}
GenHPF &Transformer &- &512 &5e-5 &4 &27.6 \\\cmidrule{1-7}
\multirow{5}{*}{Cached} &Transformer &- &4096 &1e-5 &2 &6.4 \\\cmidrule{2-7}
&Mega &- &16384 &5e-4 &2 &6.6 \\\cmidrule{2-7}
&S4 &- &16384 &1e-5 &2 &7.0 \\\cmidrule{2-7}
&Performer &- &32768 &1e-5 &2 &6.4 \\\cmidrule{1-7}
\multicolumn{2}{c}{REMed} &- &$\infty$ ($>$267k) &1e-5 &2 &6.6 \\\midrule
\bottomrule
\end{tabular}
\end{adjustbox}
\end{table*}

%% file: tables/per_task.tex
\begin{sidewaystable}[!htp]\centering
\caption{Per-task performance analysis results: bold text indicates the best performance for each task.}\label{tab:per_task}
\footnotesize
\setlength{\tabcolsep}{0.2em}
\begin{tabular}{lrrrrrrrrrr}\toprule
AUROC &Mortality\_1 &Mortality\_2 &Mortality\_3 &Mortality\_7 &Mortality\_14 &LOS\_7 &LOS\_14 &Readmission &Diagnosis \\\toprule
Flatten\_Mega &0.939 &0.904 &0.892 &0.862 &0.846 &0.817 &0.836 &0.606 &0.812 \\\cmidrule{1-10}
Flatten\_Performer &0.947 &0.911 &0.894 &0.874 &0.861 &0.817 &0.835 &0.572 &0.829 \\\cmidrule{1-10}
Flatten\_S4 &\textbf{0.949} &0.920 &0.899 &0.882 &0.864 &0.820 &0.840 &0.585 &0.831 \\\cmidrule{1-10}
GenHPF &0.935 &0.896 &0.885 &0.860 &0.845 &0.807 &0.828 &0.608 &0.830 \\\cmidrule{1-10}
Cached\_Transformer &0.868 &0.851 &0.857 &0.838 &0.829 &0.779 &0.799 &0.579 &0.824 \\\cmidrule{1-10}
Cached\_Mega &0.916 &0.888 &0.888 &0.858 &0.846 &0.793 &0.826 &0.634 &0.832 \\\cmidrule{1-10}
Cached\_Performer &0.920 &0.896 &0.883 &0.865 &0.854 &0.802 &0.833 &\textbf{0.647} &\textbf{0.844} \\\cmidrule{1-10}
Cached\_S4 &0.943 &0.921 &\textbf{0.918} &0.870 &0.856 &0.823 &0.848 &0.617 &0.832 \\\cmidrule{1-10}
REMed &0.939 &\textbf{0.921} &0.907 &\textbf{0.879} &\textbf{0.866} &\textbf{0.836} &\textbf{0.848} &0.606 &0.832 \\\toprule
AUROC &Creatinine\_1 &Creatinine\_2 &Creatinine\_3 &Platelets\_1 &Platelets\_2 &Platelets\_3 &WBC\_1 &WBC\_2 &WBC\_3 \\\toprule
Flatten\_Mega &0.902 &0.906 &0.904 &0.887 &0.898 &0.911 &0.861 &0.874 &0.872 \\\cmidrule{1-10}
Flatten\_Performer &0.910 &0.915 &0.913 &0.896 &0.909 &0.918 &0.867 &0.883 &0.881 \\\cmidrule{1-10}
Flatten\_S4 &0.961 &0.957 &0.951 &0.941 &0.939 &0.941 &0.887 &0.896 &0.892 \\\cmidrule{1-10}
GenHPF &0.962 &0.954 &0.949 &0.924 &0.927 &0.934 &0.881 &0.894 &0.890 \\\cmidrule{1-10}
Cached\_Transformer &0.967 &0.960 &0.954 &0.963 &0.957 &0.958 &0.906 &0.907 &0.906 \\\cmidrule{1-10}
Cached\_Mega &0.968 &0.960 &0.955 &0.960 &0.956 &0.957 &0.930 &0.925 &0.918 \\\cmidrule{1-10}
Cached\_Performer &0.947 &0.947 &0.943 &0.944 &0.945 &0.952 &0.898 &0.904 &0.901 \\\cmidrule{1-10}
Cached\_S4 &0.969 &0.964 &0.958 &0.952 &0.954 &0.957 &0.924 &0.918 &0.910 \\\cmidrule{1-10}
REMed &\textbf{0.987} &\textbf{0.977} &\textbf{0.971} &\textbf{0.978} &\textbf{0.971} &\textbf{0.969} &\textbf{0.961} &\textbf{0.946} &\textbf{0.936} \\\toprule
AUROC &Hemoglobin\_1 &Hemoglobin\_2 &Hemoglobin\_3 &Bicarbonate\_1 &Bicarbonate\_2 &Bicarbonate\_3 &Sodium\_1 &Sodium\_2 &Sodium\_3 \\\toprule
Flatten\_Mega &0.756 &0.749 &0.742 &0.779 &0.771 &0.775 &0.853 &0.861 &0.839 \\\cmidrule{1-10}
Flatten\_Performer &0.758 &0.755 &0.750 &0.779 &0.778 &0.777 &0.862 &0.869 &0.846 \\\cmidrule{1-10}
Flatten\_S4 &0.854 &0.831 &0.816 &0.848 &0.832 &0.813 &0.890 &0.888 &0.868 \\\cmidrule{1-10}
GenHPF &0.831 &0.809 &0.795 &0.846 &0.828 &0.815 &0.895 &0.888 &0.865 \\\cmidrule{1-10}
Cached\_Transformer &0.890 &0.865 &0.850 &0.858 &0.832 &0.824 &0.883 &0.882 &0.860 \\\cmidrule{1-10}
Cached\_Mega &0.878 &0.854 &0.841 &0.858 &0.834 &0.820 &0.906 &0.894 &0.875 \\\cmidrule{1-10}
Cached\_Performer &0.864 &0.847 &0.841 &0.832 &0.818 &0.814 &0.894 &0.889 &0.872 \\\cmidrule{1-10}
Cached\_S4 &0.893 &0.865 &0.849 &0.872 &0.848 &0.835 &0.915 &0.903 &0.882 \\\cmidrule{1-10}
REMed &\textbf{0.925} &\textbf{0.890} &\textbf{0.870} &\textbf{0.902} &\textbf{0.864} &\textbf{0.845} &\textbf{0.936} &\textbf{0.917} &\textbf{0.894} \\\midrule
\bottomrule
\end{tabular}
\end{sidewaystable}

%% file: tables/topk_human.tex
\begin{table*}[!htp]\centering
\caption{MIMIC-IV Expert-Selected Top-30 Codes}\label{ext:1}
\scriptsize
\begin{adjustbox}{width=1.0\textwidth, center}
\setlength{\tabcolsep}{0.2em}
\begin{threeparttable}
\begin{tabular}{rrr}\toprule
\textbf{Annotator 1} &\textbf{Annotator 2} &\textbf{Annotator 3} \\\cmidrule{1-3}
Skin Temperature &Heart Rate Alarm-Low &GCS-Verbal Response \\\cmidrule{1-3}
Respiratory Pattern &fspn\tnote{2} High &O$_2$ Saturation Pulseoxymetry Alarm-Low \\\cmidrule{1-3}
Breathing Pattern/Effort &Respiratory Pattern &Glucose \\\cmidrule{1-3}
Norepinephrine &Breathing Pattern/Effort &Glucose (serum) \\\cmidrule{1-3}
Respiratory Rate (total) &Norepinephrine &Respiratory Pattern \\\cmidrule{1-3}
Arterial BP\tnote{3} Diastolic &Radial Pulse L &Norepinephrine \\\cmidrule{1-3}
Inspired O$_2$ fraction &GCS\tnote{1} -Motor Response &GCS\tnote{1} -Motor Response \\\cmidrule{1-3}
Respiratory Rate &Non-Invasive BP\tnote{3} Alarm-Low &GCS\tnote{1}-Eye Opening \\\cmidrule{1-3}
Peak Insp. Pressure &Arterial BP\tnote{3} Diastolic &White Blood Cells \\\cmidrule{1-3}
Arterial BP\tnote{3} Mean &Inspired O$_2$ Fraction &Arterial BP\tnote{3} Diastolic \\\cmidrule{1-3}
Impaired Tissue Perfusion NCP\tnote{4} -Interventions &Creatinine (serum) &Creatinine (serum) \\\cmidrule{1-3}
PTT\tnote{5} &Respiratory Rate &Respiratory Rate \\\cmidrule{1-3}
Non Invasive BP\tnote{3} Mean &Arterial BP\tnote{3} Mean &Sodium \\\cmidrule{1-3}
Altered Mental Status NCP\tnote{4}-Interventions &Non Invasive BP\tnote{3} Systolic &Arterial BP\tnote{3} Mean \\\cmidrule{1-3}
Mean Airway Pressure &Arterial BP\tnote{3} Systolic &Temperature Fahrenheit \\\cmidrule{1-3}
Pupil Response Right &Non Invasive BP\tnote{3} Mean &Arterial BP\tnote{3} Systolic \\\cmidrule{1-3}
Capillary Refill R &Respiratory Effort &Pupil Response Right \\\cmidrule{1-3}
Capillary Refill L &Bicarbonate &Bicarbonate \\\cmidrule{1-3}
Tidal Volume (observed) &Respiratory Rate (spontaneous) &Heart Rhythm \\\cmidrule{1-3}
Pupil Size Right &Heart Rhythm &pH \\\cmidrule{1-3}
Heart Rate &pH &Calculated Total CO$_2$ \\\cmidrule{1-3}
HCO$_3$ (serum) &Potassium (serum) &Hemoglobin \\\cmidrule{1-3}
Resp Alarm-High &Hemoglobin &Mental Status \\\cmidrule{1-3}
spO$_2$ Desat Limit &Hematocrit (serum) &Platelet Count \\\cmidrule{1-3}
pO$_2$ &Creatinine &WBC\tnote{6} \\\cmidrule{1-3}
Heart Rate Alarm-High &pO$_2$ &Creatinine \\\cmidrule{1-3}
Anion Gap &Heart Rate Alarm-High &HCO$_3$ (serum) \\\cmidrule{1-3}
Level of Consciousness &Anion Gap &spO$_2$ Desat Limit \\\cmidrule{1-3}
Pupil Response Left &Hematocrit &Level of Consciousness \\\cmidrule{1-3}
Base Excess &Base Excess &Pupil Response Left \\\midrule
\bottomrule
\end{tabular}
\begin{tablenotes}
\setlength{\columnsep}{0.8cm}
\setlength{\multicolsep}{0cm}
{\footnotesize
\item[1] Glasgow Coma Scale
\item[2] Spontaneous Breathing Frequency
\item[3] Blood Pressure
\item[4] Nursing Care Plan
\item[5] Partial Thromboplastin Time
\item[6] White Blood Cell
}
\end{tablenotes}
\end{threeparttable}
\end{adjustbox}
\end{table*}

\begin{table*}[!htp]\centering
\caption{eICU Expert-Selected Top-30 Codes}\label{ext:2}
\scriptsize
\begin{adjustbox}{width=0.9\textwidth, center}
\setlength{\tabcolsep}{0.2em}
\begin{threeparttable}
\begin{tabular}{rrr}\toprule
\textbf{Annotator 1} &\textbf{Annotator 2} &\textbf{Annotator 3} \\\cmidrule{1-3}
Heart Rate &PTT\tnote{1} &Heart Rate \\\cmidrule{1-3}
SV\tnote{2} &Lactate &Lactate \\\cmidrule{1-3}
Norepinephrine (mcg/min) &PT\tnote{3} &Invasive BP\tnote{4} \\\cmidrule{1-3}
Non-Invasive BP\tnote{4} &Norepinephrine (mcg/min) &Pupils Right \\\cmidrule{1-3}
spO$_2$ &Invasive BP\tnote{4} &Vent Rate \\\cmidrule{1-3}
Pupils Right &spO$_2$ &Platelets x1000 \\\cmidrule{1-3}
O$_2$ Saturation &Base Deficit &HCO$_3$ \\\cmidrule{1-3}
Bicarbonate &Respiratory Assessment &Bicarbonate \\\cmidrule{1-3}
O$_2$ Sat (\%) &vitalPeriodic &Pupils Left \\\cmidrule{1-3}
Pupils Left &O$_2$ Saturation &Glasgow Coma Score \\\cmidrule{1-3}
Glasgow Coma Score &HCO$_3$ &Norepinephrine (mcg/kg/min) \\\cmidrule{1-3}
paO$_2$ &Total CO$_2$ &Pupils \\\cmidrule{1-3}
Mechanical Ventilation &Bicarbonate &pH \\\cmidrule{1-3}
Symptoms of Delirium Present &O$_2$ Sat (\%) &Vasopressin (units/min) \\\cmidrule{1-3}
Temperature &Glasgow Coma Score &Respiratory Rate \\\cmidrule{1-3}
Score (Glasgow Coma Scale) &paO$_2$ &Vasopressin (ml/hr) \\\cmidrule{1-3}
O$_2$ Content &Hct\tnote{5} &Phenylephrine (ml/hr) \\\cmidrule{1-3}
Respiratory Rate &pH &Norepinephrine (ml/hr) \\\cmidrule{1-3}
Vasopressin (ml/hr) &Respiratory Rate &WBC x1000 \\\cmidrule{1-3}
PT\tnote{3} -INR\tnote{6} &Vasopressin (ml/hr) &fiO$_2$\tnote{7} \\\cmidrule{1-3}
Norepinephrine (ml/hr) &vitalAperiodic &MAP\tnote{8} (mmhg) \\\cmidrule{1-3}
MAP\tnote{8} (mmhg) &Norepinephrine (ml/hr) &Pulse \\\cmidrule{1-3}
Capillary Refill &MAP\tnote{8} (mmhg) &BUN\tnote{9} \\\cmidrule{1-3}
paCO$_2$ &paCO$_2$ &Hgb\tnote{10} \\\cmidrule{1-3}
Pulse &Hgb\tnote{10} &Mental Status Assessment \\\cmidrule{1-3}
Mental Status Assessment &BNP\tnote{11} &Sodium \\\cmidrule{1-3}
Crystalloids &Anion Gap &Anion Gap \\\cmidrule{1-3}
Anion Gap &Arterial Line MAP\tnote{8} (mmhg) &Phenylephrine (mcg/min) \\\cmidrule{1-3}
Arterial Line MAP\tnote{8} (mmhg) &Base Excess &Arterial Line MAP\tnote{8} (mmhg) \\\cmidrule{1-3}
Base Excess &Creatinine &Creatinine \\\midrule
\bottomrule
\end{tabular}
\begin{tablenotes}
\setlength{\columnsep}{0.8cm}
\setlength{\multicolsep}{0cm}
\begin{multicols}{2}
{\footnotesize
\item[1] Partial Thromboplastin Time
\item[2] Stroke Volume
\item[3] Prothrombin Time
\item[4] Blood Pressure
\item[5] Hematocrit
\item[6] International Normalized Ratio 
\newpage
\item[7] Fraction of Inspired Oxygen
\item[8] Mean Arterial Pressure
\item[9] Blood Urea Nitrogen
\item[10] Hemoglobin
\item[11] B-type natriuretic peptide
}
\end{multicols}
\end{tablenotes}
\end{threeparttable}
\end{adjustbox}
\end{table*}